\documentclass{article}
\usepackage[T1]{fontenc}
\usepackage[utf8]{inputenc}
\usepackage{lmodern}

\usepackage{setspace}
\usepackage{geometry}
\PassOptionsToPackage{dvipsnames}{xcolor}
\PassOptionsToPackage{unicode=true,pdfusetitle,
bookmarks=true,bookmarksnumbered=false,bookmarksopen=false,
breaklinks=true,pdfborder={0 0 0},pdfborderstyle={},backref=false,colorlinks=true}{hyperref}

\usepackage{graphicx}
\usepackage{amsmath}
\usepackage{amssymb}
\usepackage{amsthm}
\usepackage{amsfonts}
\usepackage{mathrsfs}
\usepackage{mathtools}
\usepackage{appendix}
\usepackage{booktabs}
\usepackage{soul}
\usepackage{hyperref}
\usepackage{enumitem}
\usepackage{setspace}
\usepackage{float}
\usepackage{natbib}
\usepackage[ruled]{algorithm2e}
\usepackage{xcolor} 
\usepackage{stmaryrd} 

\newcommand{\argmax}{\text{argmax}}
\newcommand{\R}{\mathbb{R}}

\renewcommand{\P}{\mathbb{P}}
\newcommand{\E}{\mathbb{E}}

\newcommand{\V}{\mathbb{V}}

\newcommand{\var}{\mathrm{Var}}
\newcommand{\ind}{\mathbb{I}}
\newcommand{\indicator}{\ind}

\newcommand{\one}{\mathbf{1}}
\newcommand{\Var}{\mathrm{Var}}
\newcommand{\varp}{\mathrm{Var}^{\geq 1}}
\newcommand{\wtvar}{\widetilde{\mathrm{Var}}}

\DeclarePairedDelimiter{\parens}{(}{)}
\DeclarePairedDelimiter{\brackets}{[}{]}
\DeclarePairedDelimiter{\braces}{\{}{\}}
\DeclarePairedDelimiter{\abs}{|}{|}

\DeclarePairedDelimiter{\pospart}{(}{)_+}

\DeclarePairedDelimiter{\ceil}{\lceil}{\rceil}

\newcommand{\thin}{\,}
\newcommand{\med}{\;}
\newcommand{\wide}{\qquad}

\newcommand{\integral}{\int}

\newcommand{\eqnum}[1]{\text{°#1}}
\newcommand{\texteqnum}[1]{(\eqnum{#1})}
\newcommand{\ee}{\mathrm{e}}

\newcommand{\St}{\mathcal{S}}
\newcommand{\A}{\mathcal{A}}
\newcommand{\pistar}{\pi^\star}
\newcommand{\vstar}{V^\star}

\newcommand{\skh}{s^{k}_{h}}
\newcommand{\sko}{s^{k}_{1}}
\newcommand{\skH}{s^{k}_{H}}
\newcommand{\skhp}{s^{k}_{h+1}}

\newcommand{\akh}{a^{k}_{h}}
\newcommand{\ako}{a^{k}_{1}}
\newcommand{\akH}{a^{k}_{H}}
\newcommand{\astar}{a^{\star}}
\newcommand{\rkh}{r^{k}_{h}}
\newcommand{\rko}{r^{k}_{1}}
\newcommand{\rkH}{r^{k}_{H}}
\newcommand{\vkh}{V^{k}_{h}}
\newcommand{\vko}{V^{k}_{1}}

\newcommand{\vkhp}{V^{k}_{h+1}}
\newcommand{\vkHp}{V^{k}_{H+1}}
\newcommand{\vpiko}{V^{\pi^k}_{1}}

\newcommand{\qkh}{Q^{k}_{h}}

\newcommand{\pikh}{\pi^{k}_{h}}
\newcommand{\Pkhsa}{P^k_{h,s,a}}
\newcommand{\nkh}{n^k_h}
\newcommand{\Nkh}{N^k_h}
\newcommand{\bkh}{b^k_h}

\newcommand{\pimix}{\bar{\pi}}
\newcommand{\CBPI}{C}

\newcommand{\reg}{\mathrm{Reg}} 

\newcommand{\logp}{\log^{\geq 0}}
\newcommand{\logpp}{\log^{\geq 1}}

\usepackage{scalerel,stackengine}

\newcommand{\varstar}{\mathrm{Var}^{\star}}

\newcommand{\vardiff}{\mathrm{Var}^\mathrm{diff}}

\newcommand{\topt}{\mathscr{T}^{-}}
\newcommand{\tmodel}{\mathscr{T}^{+}}
\newcommand{\tmodelstar}{\mathscr{T}^{+}_{1}}
\newcommand{\tmodeldiff}{\mathscr{T}^+_2}
\newcommand{\tmodelbonus}{\mathscr{T}^+_3}

\newcommand{\bbar}{\bar{b}}

\newcommand{\econc}{\mathcal{E}}

\theoremstyle{plain}
\newtheorem{theorem}{\protect\theoremname}[section]
\theoremstyle{plain}
\newtheorem{lemma}[theorem]{\protect\lemmaname}
\theoremstyle{remark}

\theoremstyle{plain}
\newtheorem{corollary}[theorem]{\protect\corollaryname}
\theoremstyle{plain}

\theoremstyle{definition}

\theoremstyle{plain}
\newtheorem{proposition}[theorem]{\protect\propositionname}

\providecommand{\corollaryname}{Corollary}
\providecommand{\lemmaname}{Lemma}
\providecommand{\remarkname}{Remark}
\providecommand{\theoremname}{Theorem}
\providecommand{\conjecturename}{Conjecture}
\providecommand{\definitionname}{Definition}
\providecommand{\propositionname}{Proposition}

\usepackage[dvipsnames]{xcolor}
\usepackage[showdeletions]{color-edits} 
\addauthor[Victor]{vb}{red}
\addauthor[Matthew]{mz}{Orange}
\addauthor[Guy]{gz}{blue}
\addauthor[Yudong]{yc}{ForestGreen}
\hypersetup{
    colorlinks=true,
    linkcolor=blue,
    citecolor=blue,
}
\usepackage{subfiles}
\usepackage{microtype}
\usepackage{flafter}

\usepackage{natbib}

\hypersetup{citecolor=blue} 

\title{Closing the Log Gap: Minimax-Optimal Regret for Episodic RL}

\usepackage{authblk}
\author{Guy Zamir${}^*$}
\author{Victor Boone${}^\dagger$}
\author{Matthew Zurek${}^*$}
\author{Yudong Chen${}^*$}
\affil{
    ${}^*$Department of Computer Sciences, University of Wisconsin--Madison
    \\ ${}^\dagger$IRIT, Université de Toulouse, CNRS, Toulouse INP, Toulouse, France 
    \\\vspace{0.8em}
    \texttt{gzamir@wisc.edu}
    \quad\texttt{victor.boone@irit.fr}
    \quad\texttt{matthew.zurek@wisc.edu}
    \quad\texttt{yudongchen@cs.wisc.edu}
}

\date{}

\allowdisplaybreaks
\begin{document}

\maketitle

\begin{abstract}
    In this paper, we study the regret minimization problem in episodic tabular MDPs.
    Our algorithm achieves the first expected regret bound scaling as $\sqrt{H^3 S A K}$ for sufficiently large $K$, where $H$ is the episode length, $K$ is the number of episodes, and $S$ and $A$ are the numbers of states and actions, respectively.
    Our method is the first to match the minimax lower bound \emph{without extra polylog factors}.
    We introduce a variant of UCBVI which does not maintain strictly optimistic estimates, and uses MOSS-style exploration bonuses whose logarithmic terms adapt to both visit counts and estimated variances.
    A key challenge is controlling the propagation of errors caused by under-optimism, which we accomplish in an occupancy-weighted way, avoiding logarithmic overhead that would be incurred by uniform-over-rows concentration.
    Our analysis develops time-uniform and variance-aware concentration bounds that may be of independent interest.
    As a corollary we also develop an algorithm for $(\varepsilon, \delta)$-PAC best-policy identification, which is the first to match the $\Omega\big(H^3 SA\log(\frac{1}{\delta})\varepsilon^{-2}\big)$ minimax lower bound up to constants in some regimes.
\end{abstract}

\section{Introduction}

\begin{table*}[t!]
\centering
\resizebox{\linewidth}{!}{
    \def\arraystretch{1.33}
    \begin{tabular}{cccc}
        \textbf{Algorithm} 
        & \begin{tabular}{c}\textbf{Expected regret scaling} \\[-.33em] \textbf{as $K \to \infty$} \end{tabular} 
        & \begin{tabular}{c} \textbf{Original work is} \\[-.33em] \textbf{time-inhomogeneous} \end{tabular} 
        & \begin{tabular}{c}\textbf{Optimistic with} \\[-.33em] \textbf{high probability} \end{tabular}
        \\[.33em] \hline\hline
        \texttt{UCBVI} \cite{azar2017minimax} & $\sqrt{H^3 S A K \log(K)^2}$ & No & Yes
        \\
        \texttt{Euler} \cite{zanette2019tighter} & $ \sqrt{H^3 S A K \log(K)^6}$ & No & Yes
        \\
        \texttt{UCB-Advantage} \cite{zhang2020almost} & $ \sqrt{H^3 S A K \log(K)} + \sqrt{H^3 K \log(K)^3}$ & Yes & Yes
        \\
        \texttt{UCBMQ} \cite{menard2021ucb} & $ \sqrt{H^3 S A K \log(K)^3}$ & Yes & Yes
        \\ 
        \texttt{MVP} \cite{zhang2021reinforcement} & $ \sqrt{H^3 S A K \log(K)^2}$ & No & Yes
        \\
        \texttt{Modified MVP} \cite{zhang2025settling} & $ \sqrt{H^3 S A K \log(K)^5}$ & Yes & Yes
        \\
        \texttt{SSR} \cite{xiang2022near} & $\sqrt{H^3 S A K \log(K)^3}$ & Yes & Yes
        \\
        \texttt{OPSRL} \cite{tiapkin2022optimistic} & $\sqrt{H^3 S A K \log(K)^3}$ & Yes & Yes
        \\
        \texttt{EQO} \cite{lee2025minimax}, \texttt{EQO+} \cite{lee2026unified} & $\sqrt{H^3 S A K \log(HSA)\log(K)}$ & No & No
        \\ \hline
        \texttt{MORE-UCBVI} (ours) & $\sqrt{H^3 S A K}$ & Yes & No
        \\ \hline
        Lower Bound \cite{domingues2021episodic} & $\sqrt{H^3 S A K}$ & --- & ---
    \end{tabular}
}
\caption{
\label{table:minimax_optimal_algorithms}
    A compendium of algorithms which are minimax optimal up to at most log factors.
}
\end{table*}

In online reinforcement learning (RL), the balance of exploration and exploitation is one of the most central and investigated problems. 
A canonical setting for the study of the exploration-exploitation tradeoff is that of finite-horizon Markov decision processes (MDPs), and in particular the problem of regret minimization in such environments. 
In regret minimization, the learning agent dynamically explores the unknown environment by playing actions, and its performance is measured by comparing its aggregate reward with that of an omniscient agent that knows the environment in advance.
In this work, we are interested in the derivation of formal bounds of the worst-case regret (a.k.a.~\emph{minimax regret}) \emph{in expectation} and in the \emph{time-inhomogeneous} setting, where rewards and transitions are time-dependent. 

This problem has been extensively studied.
In the class of tabular MDPs with $S$ states, $A$ actions per state and fixed horizon $H$, it is known \citep{domingues2021episodic} that the minimax regret of every learning algorithm is at least $\smash{\Omega(\sqrt{H^3 S A K})}$, where $K$ is the total number of episodes.
In the community, \texttt{UCBVI} \citep{azar2017minimax} is recognized as the first algorithm to reach the minimax lower bound of \cite{domingues2021episodic}.
It was succeeded by many others, such as \texttt{Euler} \citep{zanette2019tighter} or \texttt{MVP} \citep{zhang2021reinforcement}, with better second order terms or better dependency with respect to the episode horizon $H$ (see Table~\ref{table:minimax_optimal_algorithms} for a more comprehensive list). 
Although these algorithms are claimed to be minimax optimal, all their regret guarantees are $\widetilde{O}(\sqrt{H^3 S A K})$, where $\widetilde{O}(-)$ is a variant of $O(-)$ that ignores logarithmic factors.
Unraveling what the $\smash{\widetilde{O}(-)}$ hides, the true regret bounds of these methods are $O(\sqrt{H^3 S A K \log(H S A K)^\gamma})$ for $\gamma > 0$ (Table~\ref{table:minimax_optimal_algorithms}).
As such, no algorithm reaches the true minimax optimal rate prior to this work.

%

In this paper, we close this longstanding gap and construct the first algorithm with minimax regret $O(\sqrt{H^3 S A K})$ for sufficiently large $K$.

\paragraph{Our contributions.}
We introduce \texttt{MORE-UCBVI}, a variant of \texttt{UCBVI} \cite{azar2017minimax} that uses \texttt{MOSS}-style bonuses \cite{audibert2009minimax} in place of \texttt{UCB}-style bonuses \cite{agrawal1995sample,auer2002finite}.

\begin{itemize}

    \item We derive a \texttt{MOSS}-style bonus for \texttt{UCBVI} whose logarithmic term adapts to both visit counts and estimated variances. The resulting algorithm (Section~\ref{sec:algorithm}, Algorithm~\ref{alg:main_alg}) deviates from the \texttt{UCBVI} framework by allowing for under-optimism; that is, its value estimates may underestimate the optimal value function.
    
    \item We obtain an expected regret bound of $O(\sqrt{H^3 S A K})$ for sufficiently large $K$ without logarithmic factors, thereby matching the minimax rate up to a universal constant for the first time (Section~\ref{sec:main_theorems}, Theorem~\ref{thm:minimax_regret}).

    \item Our main result requires a sharper analysis. We replace the standard monotonicity-based optimism induction with a stability argument that controls how under-optimism propagates through Bellman updates. This argument enables us to bound the cumulative cost of under-optimism without the logarithmic overhead induced by a union bound. We provide a more detailed proof sketch in Section~\ref{sec:proof_sketch}.

    \item As a corollary, we construct an algorithm for $(\varepsilon, \delta)$-PAC best-policy identification which is the first to match the minimax sample complexity lower bound of $\Omega(H^3 S A \log\parens{\frac 1\delta} \varepsilon^{-2})$ established by \cite{domingues2021episodic} in some regime, improving the logarithmic factor relative to prior works (Section~\ref{sec:main_theorems}, Corollary~\ref{cor:pac_bound}).

    \item Along the way, we prove a time-uniform Bernstein inequality with a linear drift, giving an analogue of \cite[Lemma~9.3]{lattimore2020bandit} for sub-gamma random variables (Section~\ref{sec:concentration}, Theorem~\ref{thm:moss_conc}). This variance-aware concentration result may be of independent interest.
\end{itemize}

\paragraph{Related work.}
The literature of regret minimization in RL is vast. 
Focusing on episodic MDPs, Table~\ref{table:minimax_optimal_algorithms} lists existing algorithms matching the minimax lower bound up to logarithmic factors.
Some are originally for the homogeneous setting, in which all $P_1, \dots, P_H$ are equal, in which case they can be applied to the inhomogeneous setting by augmenting all states with a time index in $[H]$, thus multiplying the size of the state space by $H$.
Many of these have been introduced to enrich regret minimization with parallel considerations.
\texttt{MVP} \citep{zhang2021reinforcement} is meant to improve on the dependency in the horizon $H$, for what is known as \emph{horizon-free RL}.
\texttt{EQO+} \citep{lee2025minimax,lee2026unified} provides regret guarantees in probability \emph{simultaneously} for every confidence level $\delta > 0$.
\texttt{UCBMQ} \citep{menard2021ucb} is a variant of Q-Learning \citep{watkins1992q} with near minimax optimal guarantees; it is \emph{model-free} and therefore has a cheaper memory cost.
As discussed in Section~\ref{sec:insufficiency}, the logarithmic factors incurred by these algorithms are inherent to their construction. 

Beyond the nearly minimax optimal algorithms in Table~\ref{table:minimax_optimal_algorithms}, there exists a large range of methods with formal guarantees that are interesting in their own right, such as the line of works on model-free methods \citep{menard2021ucb} or that on Bayesian methods \citep{osband2013more,tiapkin2022optimistic}. 

Furthermore, parallel to minimax regret bounds, there is a growing literature on \emph{problem-dependent} regret analysis with bounds scaling as $\log(K)$ rather than $\sqrt{K}$, see for instance \cite{xu2021fine,tirinzoni21lb,dann2021beyond,yang2021q}.
We also note that many episodic RL algorithms have analogues in \emph{average-reward RL}.
For instance, the analogue of \texttt{UCBVI} \citep{azar2017minimax} is \texttt{FOCUS} \citep{zamir2026optimal}, although the dominant template there is rather \texttt{UCRL2} \citep{auer2008near} and its variants \citep{filippi_optimism_2010,bourel_tightening_2020,boone_achieving_2024}.

\section{Preliminaries}

A \emph{tabular episodic Markov decision process} (MDP) is a tuple $(\St,\A,H,(P_h)_{h=1}^H,(r_h)_{h=1}^H,\mu)$. 
$\St$ and $\A$ are the state and action spaces with cardinalities $S$ and $A$, respectively.
$H$ is the length of each episode.
$P_h:\St\times \A \to \Delta(\St)$ is the transition kernel at step $h$, with $P_h(\cdot|s,a)$ denoting the distribution over possible next states upon taking action $a$ at state $s$.
$r_h:\St\times\A\to[0,1]$ is the (deterministic) reward function at step $h$.
$\mu$ is the initial state distribution.

We study tabular online RL.
For each episode $k=1,\dots,K$, the learner plays a policy $\pi^k = (\pikh)_{h=1}^H$ then observes the trajectory 
\(
\sko,\ako,\rko, \dots, \skH,\akH,\rkH,
\)
where $\sko\sim \mu$, then for each $h\in[H]$, $\akh= \pikh(\skh)$, $\rkh = r_h(\skh,\akh)$, and $\skhp \sim P_h(\cdot | \skh,\akh)$.
Note that we assume the learner does not know the transition kernels $(P_h)$ or initial state distribution $\mu$ but otherwise knows $\St$, $\A$, $H$, $(r_h)$, and $K$.

For a policy $\pi=(\pi_h)_{h=1}^H$ we let $(V_h^\pi)_{h=1}^{H+1}$ and $(Q_h^\pi)_{h=1}^{H+1}$ be the value and action-value functions for the policy $\pi$.
We also denote by $(V_h^\star)$ and $(Q_h^\star)$ the optimal value and action-value functions.
The learner's goal is to minimize the regret, defined as
\begin{equation}
\reg(K) \coloneq \sum_{k=1}^K \Big(\vstar_1(\sko) - \vpiko(\sko)\Big) \med .
\end{equation}
We are interested in minimizing the regret \emph{in expectation} and in the \emph{worst case}, called \emph{minimax regret}, that is, bounding the expectation of $\reg(K)$  uniformly for every MDP with $S$ states, $A$ actions and horizon $H$.

\subsection{Additional Notation}

Let $P$ be a transition kernel.
Given a step $h$ and a state-action pair $(s,a)$, we often write $P_{h,s,a}$ to denote the row vector of length $S$ whose entries are $P_h(s'|s,a)$ for each $s'$.
Hence, for $V \in\R^\St$ we use $P_{h,s,a}V$ as shorthand for $\sum_{s'}P_h(s'|s,a)V(s')$.
Given a distribution $p\in\Delta(\St)$ and value function $V\in\R^\St$, we define
\[
\var(p,V) \coloneq pV^2 - (pV)^2 = \var_{s\sim p}(V(s)).
\]
For an integer $n\geq1$, we let $[n]\coloneq\{1,\dots,n\}$.
$\one$ to denotes the all-ones vector and $\ind$ denotes the indicator function.
We write $x \lesssim y$ or $x \leq O(y)$ if $x\leq cy$ for some universal constant $c>0$, and $x\asymp y$ if both $x\lesssim y$ and $y\lesssim x$.
Finally, we define $\logp(x) \coloneq \log(\max\{1,x\})$, $\logpp(x) \coloneq \max\{1, \logp(x)$\}, $\varp(p,V) \coloneq \max\{1, \var(p,V)\}$, and $x_+\coloneq \max\{0,x\}$.

\section{Insufficiency of State-of-the-Art Methods}
\label{sec:insufficiency}

There are multiple routes to obtain an algorithm with regret of order $\sqrt{H^3 SA K \log(K)^\alpha}$.
These include variants of \texttt{UCBVI} \citep{azar2017minimax}, optimistic variants of \texttt{PSRL} \citep{osband2013more}, and optimistic variants of Q-Learning \citep{watkins1992q}.
Note that most state-of-the-art approaches are optimistic, with only a few exceptions \citep{lee2025minimax,lee2026unified}.
In this work, we take the first route and design our minimax optimal algorithm as a variant of \texttt{UCBVI}.
Our approach differs significantly from the existing family of \texttt{UCBVI}-methods,
which is perhaps unsurprising since every existing algorithm suffers from $\mathrm{polylog}(K)$ factors and is therefore not minimax optimal (Table~\ref{table:minimax_optimal_algorithms}).
In this section, we identify the culprit responsible for this suboptimality: the optimistic bonus design.

\subsection{The  \texttt{UCBVI} Algorithm}

The \texttt{UCBVI} algorithm of \cite{azar2017minimax} consists of performing dynamic programming (DP) over the MDP with overestimated reward function $(r_h + b^k_h)_{h=1}^H$ and empirical transition kernel
\begin{equation}
\label{eq:empirical_kernel}
    P^k_{h,s,a} (s')
    \coloneq
    \begin{cases}
        \frac{N_h^k (s,a,s')}{N_h^k (s,a)}
        & \text{if $N_h^k (s,a) > 0$\med,}
        \\
        \frac 1S
        & \text{if $N_h^k (s,a) = 0$\med,}
    \end{cases}
\end{equation}
where $\smash{N_h^k (s,a,s') \coloneq \sum_{i=1}^{k-1} \indicator(s_h^i = s, a_h^i = a, s_{h+1}^i = s')}$ and $N_h^k (s,a) \coloneq \sum_{s'} N_h^k (s,a,s')$ are the empirical visit counts before episode $k$.
To be more precise, given the above notation, one computes
\begin{align*}
    Q_h^k (s,a) 
    & \coloneq
    \min \begin{Bmatrix}
        r_h (s, a) + P_{h,s,a}^k V_{h+1}^k + b^k_h (s, a)
        \\
        H - h + 1
    \end{Bmatrix} \med ,
    \\
    V_h^k (s) 
    & \coloneq
    \max_{a \in \mathcal{A}} Q_h^k (s, a)
    \med ,
\end{align*}
with $V_{H+1}^k (s) \coloneq 0$ for every state $s \in \mathcal{S}$.
Then, during episode $k$, the learner plays the greedy policy with respect to $Q^k_h$:
\begin{equation*}
    \pi^k_h(s_h^k)
    \in 
    \arg \max_{a \in \mathcal{A}}
    Q_h^k (s_h^k, a)
    \med ,
\end{equation*}
breaking ties arbitrarily.

\subsection{Why \texorpdfstring{$\log(K)$}{log(K)} Appears in Variants of \texttt{UCBVI}}

To derive minimax regret bounds for \texttt{UCBVI}, the bonus  $(b_h^k)_{h=1}^H$ is designed so that $V_h^k$ upper bounds $\vstar_h$ uniformly over steps $h \in [H]$ and episodes $k \in [K]$ with high probability.
Typically, $b_h^k$ is constructed in order to invoke an empirical Bernstein inequality \citep{maurer2009empiricalbernsteinboundssample}:
\begin{equation} \label{eq:bonus_function}
    b_h^k (s, a)
    \asymp
    \sqrt{
        \frac{ \var(P_{h,s,a}^k, V_{h+1}^k)}{N_h^k (s,a)}
        \ell
    }
    + \frac{ H^2}{N_h^k (s,a)} 
    \ell
    \med , 
    \qquad\qquad
    \ell = \log \parens*{
        \frac{K}\delta
    },
\end{equation}
with the convention $b_h^k (s, a) = + \infty$ for $N_h^k (s, a) = 0$.
With this bonus, one usually obtains high-probability regret bounds of the form
\begin{equation*}
    \P \parens*{
        \!\reg(K) \!
        \ge 
        C \sqrt{
            H^3 S A K \bigl(\log(K/\delta)\bigr)^{\gamma}
        }
    }
    \!
    \le
    \delta
\end{equation*}
for some $\gamma \geq 1$.
To convert the above into a bound in expectation, one can choose the confidence parameter as $\delta \asymp 1 / \mathrm{poly}(K)$.
The reason for this choice of $\delta$ is that, when the good event in the above display fails to hold, not much can be done other than bounding $\reg(K)$ by its worst possible value $H K$.
Consequently, the expected regret is bounded by
\begin{equation*}
    \E [\reg(K)]
    \lesssim
    \sqrt{H^3SAK \bigl(\log(K/\delta)\bigr)^{\gamma}}
    + \delta \cdot H K.
\end{equation*}
This expected regret bound of order $\sqrt{H^3 S A K \bigl(\log(K)\bigr)^\gamma}$  is sub-optimal by a factor $\log(K)^{\gamma / 2}$.

\paragraph{Where the $\log(K)^{\gamma/2}$ comes from.}

The cost of optimism must at some point be paid.
In standard regret analyses, this fact appears via the sum of bonuses $\sum_{k,h} b_h^k (s_h^k, a_h^k)$.
Intuitively speaking, the regret bound suffers when the bonus is larger than necessary.
Keeping only the dominant terms of the analysis, the bound of this term goes as such:
\begin{align*}
    \sum_{k,h}
    b_h^k (s_h^k, a_h^k)
     &\overset{\eqnum{1}}
     \approx
    \sqrt{\ell} \cdot
    \sum_{k,h}
    \sqrt{ \frac{\var(P^k_{h,s_h^k,a_h^k}, \vkhp)}{N_h^k(s_h^k,a_h^k)} }
    \\ & \overset{\eqnum{2}}\approx
    \sqrt{\ell} \cdot
    \sum_{k,h}
    \sqrt{ \frac{\var(P_{h,s_h^k,a_h^k}, \vstar_{h+1})}{N_h^k(s_h^k,a_h^k)} }
    \\ & \overset{\eqnum{3}}\approx
    \sqrt{\ell} \cdot
    \sum_{s,a,h}
    \sqrt{
        N_h^{K+1} (s,a)
        \var(P_{h,s,a}, \vstar_{h+1})
    }
    \\ & \overset{\eqnum{4}}\lesssim
     \!
    \sqrt{\ell}\cdot
    \sqrt{SAH\!\sum_{s,a,h}\!
        N_h^{K+1} (s,a)
        \var(P_{h,s,a}, \vstar_{h+1})
    }
    \\ & \overset{\eqnum{5}}\lesssim
    \sqrt{\ell} 
    \cdot 
    \sqrt{H^3 S A K}
\end{align*}
where
\texteqnum{1} unfolds the definition of the bonus $(b_h^k)$;
\texteqnum{2} approximates various estimators by their true values;
\texteqnum{3} rewrites the sum over triplets $(s,a,h)$ and integrates;
\texteqnum{4} follows from Cauchy--Schwarz;
and
\texteqnum{5} follows by a well-known lemma of the field.

In the final bound after \texteqnum{5}, we recognize a term scaling as the lower bound, as well as an extra factor $\sqrt{\ell} \asymp \log(K)^{\gamma/2}$.
This factor $\sqrt{\ell}$ was pulled from the start at \texteqnum{1} and is therefore incompressible;
we pay it as soon as we have to bound $\sum_{k, h} b_h^k (s_h^k, a_h^k)$.
In other words, prior algorithms which use bonuses of the form  \eqref{eq:bonus_function} cannot attain the minimax lower bound.
We emphasize that this suboptimality is not simply an artifact of the above analysis, and that the resulting log factor in the regret is unavoidable because the fixed-log bonus is too large.
This claim is supported by lower bounds in the bandits setting for the \texttt{UCB} algorithm, which also has a fixed log factor in its bonus and is similarly doomed to incur an extra log factor in its regret \citep{kalvit2021closer}.

\section{Variance-Aware Time-Uniform Concentration}
\label{sec:concentration}

Section~\ref{sec:insufficiency} identifies the fixed logarithmic factor in standard \texttt{UCBVI} bonuses as an obstacle to achieving minimax optimality.
Now, we introduce the main concentration tool that we use to overcome this obstacle.

In stochastic bandits, \texttt{UCB} \citep{auer2002finite} was made minimax-optimal via \texttt{MOSS} \citep{audibert2009minimax}, which is essentially the same algorithm with optimistic bonuses tailored to guarantees in expectation.
Because \texttt{UCBVI} takes its roots in \texttt{UCB}, it is only natural to generalize \texttt{MOSS}'s bonuses to our setting.
The essence of \texttt{MOSS}'s bonuses is dividing by visit counts \textit{within} the logarithmic factors, replacing the static $\log(1/\delta)$ with a term of the form $\logp(1/(n\delta))$, where $n$ is an appropriate visit count.
This modification allows the logarithmic terms to shrink as observations accumulate, rather than always paying the same factor.
By taking $\delta \asymp 1/K$, it follows that the log factor vanishes when $n \asymp K$ is large.
As such and with careful analysis, the factor $\sqrt{\ell}$ from the computation \texteqnum{1--5} in Section~\ref{sec:insufficiency} becomes removable.

We now face a new challenge: these smaller bonuses need not preserve optimism.
Instead of ensuring that an estimate remains optimistic, we will control how far below the true value it can fall.
The following variance-aware, time-uniform concentration inequality provides this control.
We defer the proof to Appendix~\ref{sec:moss_conc_proof}.

\begin{theorem}[Variance-aware MOSS concentration] \label{thm:moss_conc}
Let $(X_n)_{n=1}^\infty$ be i.i.d.~random variables such that
\[
\E[X_i]=\mu,
\qquad
\var(X_i)=\sigma^2,
\qquad
|X_i-\mu|\leq c \quad\text{a.s.}
\]
For $n\geq 1$, write $\hat{\mu}_n \coloneq \frac{1}{n}\sum_{i=1}^n X_i$.
Set $\alpha_0=4$ and $\beta_0=2/3$.
For every $\delta\in(0,1)$, the random variable
\begin{equation*}
    Z_n
    \coloneq
    \mu
    - \hat{\mu}_n
    - \sqrt{\frac{\alpha_0\sigma^2}{n}\logp\parens*{\frac{1}{n\delta}}} \\
    - \frac{\beta_0 c}{n} \logp\parens*{\frac{1}{n\delta}}
\end{equation*}
satisfies $\P(\exists n \ge 1: Z_n \ge \Delta) \le \smash{16\, \delta\, \parens*{\frac{\sigma^2}{\Delta^2} + \frac c{\Delta}}}$ for every error level $\Delta > 0$.
\end{theorem}

Theorem~\ref{thm:moss_conc} is a time-uniform analogue of Bernstein's inequality with linear drift (the quantity $\Delta$).
It bounds the probability that the empirical mean plus a confidence interval underestimates the true mean by at least $\Delta$ at \textit{any} sample size.
That is, the bound applies at every sample size without a union bound.

To see how this result guides the bonus design, fix a triplet $(h,s,a)$, write $P=P_{h,s,a}$, and let $\hat P = P_{h,s,a,n}$ be its corresponding empirical transition row after $n$ observed transitions from $(h,s,a)$.
In previous works that require optimism, the bonus is designed to be an upper bound to $\bigl( P - \hat P \bigr) \vstar_{h+1}$.
Applying the usual Bernstein's inequality and then taking a union bound over $n$ leads to the fixed log factor discussed in Section~\ref{sec:insufficiency}.
Theorem~\ref{thm:moss_conc} instead suggests a bonus of the form
\[
\sqrt{\frac{\alpha_0\var\bigl( \hat P, V_{h+1}^k)}{n}\logp\parens*{\frac{1}{n\delta}}}
+ 
\frac{\beta_0 c}{n} \logp\parens*{\frac{1}{n\delta}}
\med .
\]
When $\hat P \approx P$ and $\V^k_{h+1} \approx \vstar_{h+1}$, the variance estimate is approximately $\var\bigl(P, \vstar_{h+1}\bigr)$, and Theorem~\ref{thm:moss_conc} provides control on how $\bigl( P - \hat P \bigr) \vstar_{h+1}$ relates to the bonus.
The exact bonus we use in our algorithm, which we present in the next section, has roughly the same form but requires some additional care.

\section{Our Algorithm: \texttt{MORE-UCBVI}} \label{sec:algorithm}

In this section we introduce \texttt{MORE-UCBVI} (Minimax Optimal Regret in Expectation with \texttt{UCBVI}), given in Algorithm~\ref{alg:main_alg}.
We build on the \texttt{UCBVI} framework described in Section~\ref{sec:insufficiency}.
To recap, before each episode $k$ we update an empirical transition kernel according to \eqref{eq:empirical_kernel} then use it in value iteration to compute our value estimates $Q^k$ and $V^k$.
Then for the duration of the episode we act greedily with respect to $Q^k$, with ties between best actions broken arbitrarily.

\begin{algorithm}
\caption{
    \texttt{MORE-UCBVI}: \textbf{M}inimax \textbf{O}ptimal \textbf{R}egret in \textbf{E}xpectation for \texttt{UCBVI}
} 
\label{alg:main_alg}
\SetAlgoLined
\DontPrintSemicolon

\For{$k=1,\dots,K$}{

    Compute empirical kernels $P^k$, see Eq.~\eqref{eq:empirical_kernel}
    
    Set $\vkHp(s) \gets 0$ for all states $s$ \;

    \For(\tcp*[f]{optimistic DP}){$h=H,\dots,1$}{
        $\qkh(s,a) \gets \min \bigl\{ r_h(s,a) + \Pkhsa\vkhp + \bkh(s,a), H-h+1 \bigr\}$ for all pairs $(s,a)$ \;
        $\vkh(s) \gets \max_{a} \qkh(s,a)$ for all states $s$ \;
    }

    Observe the initial state $\sko$ \;

    \For(\tcp*[f]{run episode}){$h=1,\dots,H$}{
        Take action $\akh = \argmax_a \qkh(\skh,a)$ \;
        Observe the new state $\skhp$ \;
    }
}
\end{algorithm}

The difference between the original \texttt{UCBVI} of \cite{azar2017minimax} and our algorithm lies in the bonus.
Write $\nkh(s,a) \coloneq \max\bigl\{ 1, \Nkh(s,a) \bigr\}$, $B \coloneq HK/(SA)$.
For a distribution $p\in\Delta(\St)$, visit count $n\geq 1$, and value function $V \in\R^\St$, we define a bonus function
\begin{equation}
\label{eq:bonus_function_final}
    b(p,n,V) \coloneq 
    \max \braces[\Big]{
        b_1 (p, n, V),
        b_2 (p, n, V)
    }
\end{equation}
where
\begin{align}
    b_1 (p, n, V)
\nonumber
    & \coloneq
    \sqrt{
        \frac{\alpha \wtvar(p, n, V)}n
        \logpp \! \parens*{
            \frac{B}{ n \wtvar(p, n, V)}
        }
    }
    \\ b_2 (p, n, V)
\nonumber
    & \coloneq 
    \frac{\beta H^2S L}{n}
    \quad \text{with} \quad L \coloneq \log (10HS^2 A K^4)
    \med .
\end{align}
Here, $\alpha=64$, $\beta = 4096$, $B=HK/(SA)$, and $\wtvar$ is a high probability~upper-bound of the variance:
\[
    \wtvar(p,n,V) 
    \coloneq 
    2\thin\varp(p,V) + \frac{64H^2SL}{n}
    \med .
\]
Then the bonus used by the algorithm to compute $\qkh(s,a)$ before episode $k$ is simply the bonus function evaluated at the corresponding empirical transition kernel, count, and next-step value function:
\[
    \bkh(s,a)
    \coloneq
    b\parens*{P^k_{h,s,a}, \nkh(s,a), \vkhp}
    \med .
\]
We remark that $B$ can be interpreted as an exploration budget because the logarithmic factor decreases to 1 as the visit count increases to $B/\wtvar$.
Choosing larger $B$ results in larger bonuses, while decreasing $B$ yields worse under-optimism.
The choice $B=HK/(SA)$ optimally balances this tradeoff.
optimize $\alpha$ and $\beta$, which could likely be significantly improved.

The bonus $b \coloneq \max \braces{b_1, b_2}$ is typical of sub-gamma random variables.
The part $b_1$ is due to a sub-gaussian tail that dominates for large $n$, while $b_2$ is the shadow of a sub-exponential tail that outweighs $b_1$ for small $n$.
Note that $b_1$ is a \texttt{MOSS}-style bonus while $b_2$ is reminiscent of a \texttt{UCB}-style bonus.
Because $b_1$ ends up being dominant for large $n$, taking it in \texttt{MOSS}-style is crucial to have tight guarantees in expectation.
On the other hand, taking $b_2$ as \texttt{UCB}-style instead will be more appropriate for the use of good events.
In addition, note the presence of an optimistic estimator of variance in $b_1$.
Its role cannot be explained by Theorem~\ref{thm:moss_conc} alone, so we defer further clarifying its presence to Section~\ref{sec:over_optimism}.

\section{Main Theorem}
\label{sec:main_theorems}

We now present the main result of our paper.

\begin{theorem} \label{thm:minimax_regret}
    The regret of Algorithm~\ref{alg:main_alg} satisfies
    \[
    \E\left[ \reg(K) \right]
    \leq
    O\left( \sqrt{H^3SAK} + H^3S^2A(\log K)^2 \right).
    \]
\end{theorem}

We provide a proof sketch in Section~\ref{sec:proof_sketch} and a complete proof in Appendix~\ref{sec:complete_proof}.
The leading term in Theorem~\ref{thm:minimax_regret} matches the $\Omega\bigl(\sqrt{H^3SAK}\bigr)$ minimax lower bound of \cite{domingues2021episodic} for time-inhomogeneous episodic MDPs up to a universal constant.
Unlike the results of previous works summarized in Table~\ref{table:minimax_optimal_algorithms}, the leading term of our regret bound contains no extra logarithmic factors.
Theorem~\ref{thm:minimax_regret} is therefore the first result to completely close the gap between upper and lower bounds (for $K$ greater than a polynomial of $H, S, A$).

\paragraph{An application to best policy identification.}
We next provide an application of our main result to the problem of best-policy identification \citep{domingues2021episodic}. We say that an algorithm is $(\varepsilon, \delta)$-PAC if, for every MDP, it terminates after a finite number of episodes and returns a policy $\hat{\pi} = (\hat{\pi}^h)_{h=1}^H$ satisfying
\begin{equation*}
    \P \left(\vstar_1(\mu) - V_1^{\hat{\pi}}(\mu) \leq \varepsilon \right) \geq 1 - \delta
    \med .
\end{equation*}
Here $V_1^{\hat{\pi}}(\mu)=\sum_s \mu(s)V_1^{\hat{\pi}}(s)$ is the expected value function with respect to the starting distribution $\mu$, and likewise $\vstar_1(\mu) = \sum_s \mu(s)\vstar_1(s)$.
\begin{corollary} \label{cor:pac_bound}
    There exists an algorithm which, for any $\varepsilon \in (0,H], \delta \in (0,1]$, is $(\varepsilon, \delta)$-PAC for best-policy identification and uses at most
    \begin{equation*}
        O\left( \frac{H^3 SA L_\delta}{\varepsilon^2} + \frac{H^2 L_\delta^2}{\varepsilon^2} + H^3S^3 A (\log(eHSA))^4 L_\delta \right)
    \end{equation*}
    episodes, where $L_\delta = \log(2/\delta)$.
\end{corollary}
In the regime where $\varepsilon$ is sufficiently small and $\delta$ is not exponentially small in $SAH$, specifically $\varepsilon \leq \frac{1}{S (\log (eHSA))^2}$ and $\delta \geq \exp(-SAH)$, the total number of episodes used is
$O\bigl(H^3 SA \log(\frac {2}{\delta}) \varepsilon^{-2}\bigr)$.
To the best of our knowledge this is the first result to match the 
$\Omega\bigl(H^3 S A \log(\frac 1\delta) \varepsilon^{-2}\bigr)$
minimax lower bound shown by \citet{domingues2021episodic} in some regimes, improving the logarithmic factors relative to prior work.

Corollary \ref{cor:pac_bound} follows easily from Theorem \ref{thm:minimax_regret} by applying standard regret-to-PAC techniques \citep{kaufmann2021adaptive}: we can run Algorithm \ref{alg:main_alg} for sufficiently many episodes $K_0$ so that $\E [\reg(K_0)/K_0] \leq O(\varepsilon)$, use Markov's inequality to conclude that a uniformly randomly selected policy from $\pi_1, \dots, \pi_{K_0}$ has suboptimality $O(\varepsilon)$ with at least constant probability, and then boost the success probability by repeating $O(\log (1/\delta))$ times, evaluating all policies to error $O(\varepsilon)$ and returning the best one.
We refer to Appendix \ref{sec:BPI_appendix} for the formal algorithm statement, proof, and more discussion of related work on best-policy identification and the regimes where the lower bound of \cite{domingues2021episodic} holds.

\section{Proof Sketch}
\label{sec:proof_sketch}

In this section we sketch the proof of Theorem~\ref{thm:minimax_regret}, focusing on the main technical contributions.
We defer the full analysis to Appendix~\ref{sec:complete_proof}.

Using standard arguments (see \cite{azar2017minimax, zhang2021reinforcement} and Lemma~\ref{lem:regret_decomposition}), we decompose the expected regret into two 
\begin{equation}\label{eq:main_regret_decomposition}
    \E[\reg(K)] \leq \E[\topt] + \E[\tmodel],    
\end{equation}
where
\[
    \topt \coloneq \sum_{k} \left( \vstar_1(\sko) - \vko(\sko) \right)_+,
    \qquad
    \tmodel \coloneq \sum_{k,h} \Bigl( \left(P^k_{h,\skh,\akh} - P_{h,\skh,\akh}\right) \! \vkhp + \bkh(\skh,\akh) \Bigr)_+.
\]  
The term $\topt$ quantifies the total cost due to the optimistic value $\vko(\sko)$ underestimating the true value $\vstar_1 (\sko)$, and is therefore a \emph{cost of under-optimism}.
That is, as the bonus decreases and the algorithm becomes less optimistic, $\topt$ increases.
Meanwhile, $\tmodel$ quantifies how much the optimistic value $\vko(\sko)$ tends to overshoot the true value $\vstar_1 (\sko)$, and is therefore a \emph{cost of over-optimism}; the larger the bonus, the larger $\tmodel$.
To have the best regret possible, these two terms must be balanced.
In particular, and in opposition to past analysis such as \cite{azar2017minimax} and \cite{zhang2021reinforcement}, we will see that the \texttt{MOSS}-style bonuses no longer ensure that $\topt$ vanishes.

\subsection{Bounding \texorpdfstring{$\topt$}{T-}: Under-Optimism}

As mentioned in Section~\ref{sec:insufficiency}, optimistic algorithms choose bonuses large enough to ensure that $\vkh \geq \vstar_h$ simultaneously for all $k,h$ -- and therefore $\topt \leq 0$ -- with high probability.
We first recall the role of \emph{monotonicity} in this argument before explaining what changes for our smaller MOSS-style bonus.

\paragraph{The argument of \texttt{UCBVI} and its successors.}

Consider a \texttt{UCB}-style bonus $b_\mathrm{ucb}$ for which the Bellman update $V \mapsto pV + b_\mathrm{ucb}(p,n,V)$ is monotone, as in the analysis of \texttt{MVP} \citep{zhang2021reinforcement}.
Fix $k$, and let $V^k_h$, $h\in[H+1]$, denote the value estimates computed before the $k$-th episode using \texttt{UCBVI} with the bonus $b_\mathrm{ucb}$.
The proof shows $\vkh \geq \vstar_h \,\,\forall h$ by backwards induction over $h$.
The base case is of course trivial because $\vkHp = \vstar_{H+1}=0$.
Then, for the inductive case, suppose that $\smash{\vkhp \geq \vstar_{h+1}}$ and let $s$ be arbitrary.

For ease of notation, write $\astar = \pistar_h(s)$, $P = P_{h,s,\astar}$, $\hat{P} = P^k_{h,s,\astar}$ and $n = \nkh(s,\astar)$.
If $\vkh(s) = H-h+1$, the claim follows immediately.
Otherwise, with high probability,
\begin{align*}
    \vkh(s) 
    & \overset{\eqnum{1}}\geq r_h(s,\astar) + \hat{P} \vkhp + b_\mathrm{ucb}\bigl(\hat P,n,\vkhp\bigr) \\
    & \overset{\eqnum{2}}\geq r_h(s,\astar) + \hat{P} \vstar_{h+1} + b_\mathrm{ucb}\bigl(\hat P,n,\vstar_{h+1}\bigr) \\
    & \overset{\eqnum{3}}\geq r_h(s,\astar) + P \vstar_{h+1} = \vstar_h(s).
\end{align*}
Here, the inequality \texteqnum{1} is by definition of $\vkh$, and the inequality \texteqnum{3} is because $b_\mathrm{ucb}$ is large enough so that 
\begin{equation} \label{eq:sketch_opt_conc}
\big( P - \hat P)\vstar_{h+1} \leq b_\mathrm{ucb}\bigl(\hat{P},n,\vstar_{h+1}\bigr)
\end{equation}
with high probability (simultaneously for all $h,s,a,n$).
The inequality \texteqnum{2} crucially depends on the aforementioned monotonicity of the Bellman update.
In particular, $\vkhp \geq \vstar_{h+1}$ implies $\hat{P}\vkhp + b_\mathrm{ucb}(\hat P,n,\vkhp) \geq \hat{P}\vstar_{h+1} + b_\mathrm{ucb}(\hat P,n,\vstar_{h+1})$.
This monotonicity allows us to replace the data-dependent vector $\vkhp$ by the fixed vector $\vstar_{h+1}$ before applying concentration, thereby avoiding potentially loose covering arguments over value functions.

\paragraph{Controlling the cost of under-optimism.}
For our smaller \texttt{MOSS}-style bonus $b$, a concentration inequality of the form of Eq.~\eqref{eq:sketch_opt_conc} need not hold simultaneously for all $h,s,a,n$ with high enough probability.
We therefore cannot assume $\vkhp \geq \vstar_{h+1}$ at the next step of the induction, even though the function $V \mapsto \hat P V + b(\hat P,n,V)$ is still monotone (see Corollary~\ref{cor:monotonicity}).
However, while we cannot order $\vkhp$ and $\vstar_{h+1}$ in a strict manner, we can instead control how a next-step shortfall propagates through the Bellman update.
This is made possible by the following result, which is a slightly simplified version of Corollary~\ref{cor:stability}.

\begin{lemma}[Stability of the Bellman Update]
\label{lem:sketch_bonus_stability}
    Let $p\in\Delta(\St)$, $n\geq 1$, and $V,V'\in[0,H]^\St$. Then
    \[
        pV + b(p,n,V)
        \geq
        pV' + b(p,n,V') - \left(1 + \frac{1}{16H} \right) p(V' - V)_+ \med .
    \]
\end{lemma}

Applying this lemma with $V=\vkhp$ and $V'=\vstar_{h+1}$ gives nearly the same substitution as monotonicity, except that now $\vkhp$ and $\vstar_{h+1}$ may be unordered.
Moreover, this substitution only incurs an extra cost of approximately $\hat P (\vstar_{h+1} - \vkhp)_+$.
Roughly speaking, the previous argument with $b_\mathrm{ucb}$ was that ``$\vkhp$ is optimistic so $\vkh$ is optimistic''.
The new argument with $b$ states that ``$\smash{\vkhp}$ is nearly optimistic so $\smash{\vkh}$ is nearly optimistic''.

Explicitly, under a certain high probability good event $\econc$, stability yields
\[
    \pospart{\vstar_h(s) - \vkh(s)}
    \leq {} 
    \pospart*{
        \parens{P - \hat P} \vstar_{h+1} 
        \! - \bbar\bigl(\hat P, n, \vstar_{h+1} \bigr)
    } 
     + \Bigl( 1 + \frac{1}{4H} \Bigr) P \pospart*{\vstar_{h+1} - \vkhp}
    \med .
\]
There are three notable adjustments which require explaining:
instead of $b$ we have $\bbar$; instead of $(1 + 1/(16H))$ we have $(1 + 1/(4H))$; and instead of $\hat P$ we have $P$.
To obtain the desired bound on $(\vstar_1(s) - \vko(s))_+$ we will unroll the above inequality backwards from $h=H+1$ to 1, and for reasons that will shortly be apparent, it is important that the next-step shortfall is weighted by $P$ instead of $\hat P$.
We accomplish this change by splitting off half the bonus and using it to approximately upper bound $\bigl| \hat P - P \bigr|$ under the good event $\econc$.
$\bbar = b/2$ is the remaining half of the bonus, and the ``approximate'' upper bounding is what necessitates changing $(1 + 1/(16H))$ to $(1 + 1/(4H))$.

Since $\vkHp = \vstar_{H+1} = 0$, as mentioned above, we can unroll the previous inequality backwards.
Using $(1 + 1/(4H))^H \leq 2$, what remains is an occupancy weighted sum of only the first terms.
Specifically,
\[
    \mathbb E\bigl[\mathbb I(\mathcal E)(V_1^\star(\sko)-V_1^k(\sko))_+\bigr] 
    \lesssim
    \sum_{h,s,a}
    \P_{\mu}^{\pi^\star} 
    \begin{pmatrix} 
        s_h = s \\ a_h = a 
    \end{pmatrix}
    \E \Bigl[
        \indicator(\mathcal{E})
        \Bigl( (P_t - P_t^k) \vstar_{h+1} - \bbar(P_t^k, n_t^k, \vstar_{h+1}) \Bigr)
    \Bigr].
\]
Here, $\P^{\pistar}_\mu(s_h=s,a_h=a)$ is the probability of visiting $(s,a)$ at step $h$ in the true MDP, starting from a state randomly drawn from $\mu$ then following $\pistar$.
These occupancies belong to the true MDP under $\pistar$ and are thus deterministic.
Consequently, we only need an expected shortfall bound for each fixed $(h,s,a)$, rather than a simultaneous guarantee for all $(h,s,a)$.
Note that if we kept $\hat P$ instead of $P$, we would encounter issues due to data-dependent occupancies.

We obtain the following lemma by applying the time-uniform  concentration inequality Theorem~\ref{thm:moss_conc} and integrating its tail bound. Time-uniformity is essential because $\nkh(s,a)$ is chosen adaptively, and we cannot afford to take a union bound over its possible values.
\begin{lemma}
For each episode $k$\! and triplet $t \!\equiv \! (h,s,a)$,
\[
    \E \brackets*{
        \ind (\econc) 
        \pospart*{
            (P_t - P^k_t) \vstar_{h+1}
            - \bbar(P^k_t, n^k_t, \vstar_{h+1})
        }
    }
    \lesssim 
    \frac{1}{\sqrt{B}}\varp(P_t, \vstar_{h+1})
    \med .
\]
\end{lemma}
All that remains is accounting for $\econc^c$, a standard argument to bound a sum of variance terms, summing over $K$, and plugging in $B = HK/(SA)$.
These imply
\[
\E[\topt]
\lesssim
\frac{KH^2}{\sqrt{B}} + \frac{H}{K^2}
\lesssim
\sqrt{H^3SAK}.
\]

\subsection{Bounding \texorpdfstring{$\tmodel$}{T+}: Over-Optimism}
\label{sec:over_optimism}

To bound the over-optimism error $\tmodel$, we utilize the splitting $\vkhp = \vstar_{h+1} + ( \vkhp - \vstar_{h+1})$.
Specifically, we upper bound $\tmodel$ by the sum of three terms:
\begin{align*}
    \tmodelstar &= \sum_{k,h} \left( \left( P^k_{h,\skh,\akh} - P_{h,\skh,\akh} \right) \vstar_{h+1} \right)_+, \\
    \tmodeldiff &= \sum_{k,h} \left|\left( P^k_{h,\skh,\akh} - P_{h,\skh,\akh} \right) \left( \vkhp - \vstar_{h+1} \right) \right|, \\
    \tmodelbonus &= \sum_{k,h} \bkh(\skh,\akh) \med .
\end{align*}
By splitting like so, we are allowed to apply concentration results to the fixed vector $\smash{\vstar_{h+1}}$, thereby avoiding in $\tmodelstar$ issues of dependence between the empirical transition kernel $\smash{P^k_{h,\skh,\akh}}$ and the value estimate $\vkhp$.

Overall, the decomposition of $\tmodelstar, \tmodeldiff, \tmodelbonus$ is standard and can be found in \cite{zhang2020almost} for instance.
We can handle $\tmodeldiff$ using a standard variance analysis along the learner's trajectory.
For the other sums $\tmodelstar$ and $\tmodelbonus$, however, we must develop sharper arguments to avoid extra logarithms in the leading term.
This is what we discuss in the remainder of the section.
The detailed analyses of $\E[\tmodelstar]$, $\E[\tmodeldiff]$, and $\E[\tmodelbonus]$ are found in Appendix~\ref{sec:appendix_amount_over_optimism}.

\paragraph{Bounding $\tmodelstar$: Concentration of cumulative deviations.}
For a fixed triplet $(h,s,a)$, bounding $d_n \coloneq \bigl(\hat{P}_{h,s,a,n} - P_{h,s,a}\bigr)\vstar_{h+1}$ simultaneously for all $n\in[K]$ would require a union bound over $K$ events, incurring a factor of $\log(K)$, or at least $\log\log(K)$ by the law of iterated logarithms.
Instead, we develop a general result  --- which may be of independent interest --- to directly bound the sum of deviations $\sum_n d_n$ (proved in Appendix~\ref{sec:appendix_amount_over_optimism}, Proposition~\ref{prop:sum_slack_conc}).

\begin{proposition}[Concentration on sums with slackness]
\label{prop:sketch_sum_slack}
Let $(X_n)_{n=1}^\infty$ be i.i.d.\ random variables with mean $\mu$, variance $\sigma^2$, and $|X_i-\mu|\leq c$ almost surely. Write $\hat\mu_n=n^{-1}\sum_{i=1}^nX_i$ the empirical mean. There is a universal constant $C>0$ such that, for every $\Delta>0$,
\[
\sum_{n=1}^{\infty}\E[(\hat\mu_n-\mu-\Delta)_+]
\leq C\left(\frac{\sigma^2}{\Delta}+c\right).
\]
\end{proposition}

The following corollary helps bound $\tmodelstar$.
It follows by adding and subtracting a slackness parameter $\Delta$, then tuning to the optimal choice $\Delta = \sqrt{\sigma^2/\E[N]}$.

\begin{corollary}[Concentration on sums]
\label{cor:sketch_sum}
Under the setting of Proposition~\ref{prop:sketch_sum_slack}, for any nonnegative integer-valued random variable $N$ with $\E[N]<\infty$,
\[
\E\!\left[\sum_{n=1}^N(\hat\mu_n-\mu)_+\right]
\lesssim\sqrt{\sigma^2\E[N]}+c.
\]
\end{corollary}

Recall that $N_h(s,a)$ is the total number of visits to $(h,s,a)$ over the $K$ episodes.
After rearranging the sum over $k,h$ into a sum over $h,s,a,n$, Corollary~\ref{cor:sketch_sum} gives
\[
    \E \brackets*{
        \ind (\econc) 
        \pospart*{
            (P_t - P^k_t) \vstar_{h+1}
            - \bbar(P^k_t, n^k_t, \vstar_{h+1})
        }
    }
    \lesssim 
    \frac{1}{\sqrt{B}}\varp(P_t, \vstar_{h+1})
    \med .
\]
where the second inequality follows from Cauchy-Schwarz and Lemma~\ref{lem:var_sum}.

\paragraph{Bounding $\E[\tmodelbonus]$: Summing the bonus.}

Up to lower order terms incurred by converting from empirical estimates to true values, we have 
\[ 
    \bkh(s,a) 
    \approx 
    \sqrt{
        \frac{\sigma_h (s,a)^2}{\nkh(s,a)}
        \logpp \parens*{
            \frac{B}{\nkh(s,a) \thin \sigma_{h}(s,a)^2}
        }
    } 
\]
where $\sigma_{h}(s,a)^2 \coloneq \varp(P_{h,s,a},\vstar_{h+1})$ is the local transition variance.
Hence, $\E[\tmodelbonus]$ is approximately
\begin{equation} 
\label{eq:bonus_sum_sketch}
\sum_{h,s,a} 
\E\brackets*{
    \sum_{n=1}^{N_h(s,a)} 
    \sqrt{
        \frac{\sigma_h (s,a)^2}n
        \logpp \parens*{
            \frac {B}{ n \thin \sigma_h (s,a)^2 }
        }
    }
}
\end{equation}
Unlike the fixed logarithmic factor in the bonus in Section~\ref{sec:insufficiency}, here the logarithmic term $\logpp(B/( n \thin \sigma_{h}(s,a)^2))$ decreases as the visit count $n$ increases.
Crucially, we compute the sum with the decreasing logarithmic terms directly, rather than upper bounding all logs by their largest value and pulling a $\sqrt{\log B}$ factor outside the sum.
We formalize this in the following lemma.

\begin{lemma}[Variance-normalized MOSS sum]
\label{lem:sketch_moss_sum}
For every $q>0$, $B>0$, and integer $N\geq0$,
\[
    \sum_{n=1}^N
    \sqrt{
        \frac qn
        \logpp \parens*{
            \frac {B}{ n q}
        }
    }
    \leq
    6\sqrt{B}+ \frac{q N}{\sqrt{B}}
    \med .
\]
\end{lemma}

We remark that, as opposed to \texttt{UCB}-style bonuses, summing \texttt{MOSS}-style bonuses via Lemma~\ref{lem:sketch_moss_sum} does not spawn logarithmic terms.
The proof splits the sum according to $n\leq B/q$ and $n>B/q$.
For $n\leq B/q$, an integral comparison sums the bonuses to at most $6\sqrt{B}$.
For $n>B/q$, the logarithmic factor equals $1$, so each bonus is at most $q / \sqrt{B}$, giving the remaining $q N /\sqrt{B}$.

With all the results so far, we can finally explain when the inclusion of the variance in the denominator of the logarithm of $b_1$ in Eq.~\eqref{eq:bonus_function_final} makes its play.
Without it, the same integral comparison as before would yield a first term of $\sum_{h,s,a}\sqrt{\sigma_{h}(s,a)^2 B }$, whose worst-case bound is of order $\smash{H^2SA\sqrt{B}=\sqrt{H^5SAK}}$, a factor $H$ larger than the desired $\sqrt{H^3SAK}$ rate. 
Including the variance in the denominator changes the range of the first sum from $n\leq B$ to $n\leq B/q$.
The integral comparison then gives $\sqrt{B}$ rather than $\sqrt{qB}$, since the factor $\sqrt{q}$ in the bonus is exactly canceled by the factor $1/\sqrt{q}$ from the shorter summation range. 

Applying Lemma~\ref{lem:sketch_moss_sum}, we get
\begin{align*}
    \eqref{eq:bonus_sum_sketch} 
    &\lesssim HSA\sqrt{B} + \frac{1}{\sqrt{B}} \sum_{h,s,a} \sigma_{h}(s,a)^2\E[N_h(s,a)] \\
    &\lesssim HSA \sqrt{B}+ \frac{{H^2K}}{\sqrt{B}}
    \lesssim \sqrt{H^3SAK}.
\end{align*}
The last line follows from a folklore result on the variance sum (Lemma~\ref{lem:var_sum}), then plugging in $B = \frac{HK}{SA}$.

\section{Conclusion}

In this paper we match the minimax regret lower bound in episodic RL without extra polylogarithmic factors, using a MOSS-style bonus which incorporates visit counts and variance estimates. We believe that these algorithmic techniques, as well as our concentration results and analysis of the resulting under-optimism, may be more widely useful.
Future directions include improving the additive ``burn-in'' regret term (similar to \cite{zhang2025settling}) and handling the time-homogeneous setting.

\section*{AI Acknowledgment}

The problem formulation and most of the proof ideas were developed by the authors. The initial writing was completely done by the authors. ChatGPT 5.6 Sol and ChatGPT 6 Astra were used for the following: (i) they helped refine some of our arguments, including removing a $\log H$ factor by inserting the variance approximation $\wtvar$ in the denominator of the log in the bonus, and it helped to remove a $\log S$ factor with the stability lemma; (ii) they helped with literature review, catching typos, improving the writing, and double-checking the correctness of proofs.

Any components which were assisted with LLMs were thoroughly reviewed and re-written with our own notation, exposition, and understanding.
We take complete responsibility for the final content of this work, including text, claims, or artifacts produced with the aid of LLMs.

\section*{Acknowledgment}

G.\ Zamir, M.\ Zurek, and Y.\ Chen acknowledge support by National Science Foundation grants CCF-2233152
and DMS-2023239, by a Cisco Systems Fellowship, and by a Vilas Associates Award.

\bibliography{refs}

\clearpage

\clearpage
\appendix
\thispagestyle{empty}

\onecolumn

\numberwithin{equation}{section}
\numberwithin{theorem}{section}
\clearpage

\section{Complete Proof of Theorem~\ref{thm:minimax_regret}} \label{sec:complete_proof}

In this section we prove our main result, Theorem~\ref{thm:minimax_regret}.

\subsection{Overview of Notation and Conventions}

We recall notation used throughout the proof.
Unless stated otherwise, sums over $k$ and $h$ range over $[K]$ and $[H]$, respectively, and inequalities between vectors are understood componentwise.
For each $(h,s,a)\in[H]\times\St\times\A$, let $(X_{h,s,a,i})_{i\geq1}$ be an i.i.d.~sequence with law $P_{h,s,a}$.
We take these sequences to be mutually independent and independent of the initial states $(\sko)_{k=1}^K$, which are drawn i.i.d.\ from $\mu$.
Equivalently, the trajectory can be generated by revealing the next unused sample from the corresponding sequence whenever the learner visits $(h,s,a)$.
For every integer $n\geq1$, define the empirical row
\begin{equation}
\label{eq:kernel_empirical_row}
    \hat P_{h,s,a,n}(s')
    \coloneq
    \frac1n\sum_{i=1}^n\ind\{X_{h,s,a,i}=s'\},
    \qquad s'\in\St.
\end{equation}
This definition also applies to sample sizes that the learner does not reach.

The following table summarizes the transition and count notation:
\begin{center}
\renewcommand{\arraystretch}{1.2}
\begin{tabular}{@{}lp{0.63\linewidth}@{}}
\hline
Notation & Meaning \\
\hline
$P_{h,s,a}$ & True transition row at $(h,s,a)$. \\
$P^k_{h,s,a}$ & Transition row used by the algorithm in episode $k$, as defined in Eq.~\eqref{eq:empirical_kernel}. \\
$\hat P_{h,s,a,n}$ & Empirical row after the first $n$ samples from $(h,s,a)$, see Eq.~\eqref{eq:kernel_empirical_row}. \\
$\Nkh(s,a)$ & Number of visits to $(h,s,a)$ before episode $k$. \\
$\nkh(s,a)$ & Safeguarded count $\max\{1,\Nkh(s,a)\}$. \\
$N_h(s,a)$ & Total number of visits to $(h,s,a)$ after all $K$ episodes. \\
\hline
\end{tabular}
\end{center}
In particular,
\[
    \Nkh(s,a)
    \coloneq\sum_{i=1}^{k-1}\ind\{s_h^i=s,\ a_h^i=a\},
    \qquad
    N_h(s,a)
    \coloneq\sum_{k=1}^K\ind\{\skh=s,\ \akh=a\}.
\]
When $\Nkh(s,a)\geq1$, we have
$P^k_{h,s,a}=\hat P_{h,s,a,\Nkh(s,a)}=\hat P_{h,s,a,\nkh(s,a)}$.
When $\Nkh(s,a)=0$, the algorithm instead uses the uniform row $P^k_{h,s,a}(s')=1/S$.
We do not identify this initialization with $\hat P_{h,s,a,1}$, so we handle first visits separately when using concentration bounds for the empirical rows.

We also recall that $B=HK/(SA)$ and $L=\log(10 HS^2AK^4)$, and that
\[
    \logp(x) \coloneq \log(\max\{1,x\}),
    \qquad
    \logpp(x) \coloneq \max\{1,\logp(x)\},
    \qquad
    \varp(p,V) \coloneq \max\{1,\var(p,V)\}.
\]
Powers and positive parts of vectors are taken componentwise.

The algorithm uses the bonus $\bkh(s,a) = b\bigl(P^k_{h,s,a},\nkh(s,a),\vkhp)$.
The bonus function $b$ is defined by $b= \max\{b_1,b_2\}$, where $b_1$ and $b_2$ are defined as in Section~\ref{sec:algorithm}.
For the under-optimism analysis we define the half-bonus
\[
    \bbar (p,n,V)\coloneq\frac12 b(p,n,V).
\]
As briefly discussed in Section~\ref{sec:proof_sketch}, we will use $\bbar$ to compare value functions and reserve the remaining half of $b$ to replace empirical transitions by true transitions.
Finally, we recall our choices of constants within the bonus:
\[
\alpha \coloneq 64, \qquad \beta \coloneq 4096.
\]

\subsection{Regret Decomposition}

Define
\[
\tmodel \coloneq \sum_{k,h} \Bigl( \bigl(P^k_{h,\skh,\akh} - P_{h,\skh,\akh}\bigr)\vkhp + \bkh(\skh,\akh) \Bigr)_+
, \qquad\quad
\topt \coloneq \sum_{k} \Big( \vstar_1(\sko) - \vko(\sko) \Big)_+.
\]
With the following lemma, we split the regret into the sum of these two terms.
We defer the proof to Section~\ref{sec:regret_decomposition_proof}.
\begin{lemma}[Regret Decomposition, Eq.~\eqref{eq:main_regret_decomposition}] \label{lem:regret_decomposition}
We have
\[
    \E\left[ \reg(K) \right] \leq  \E[\topt] + \E[\tmodel].
\]
\end{lemma}
As discussed in Section~\ref{sec:proof_sketch}, we view $\topt$ as the cost of under-optimism and $\tmodel$ as the cost of over-optimism.

\subsection{High Probability Good Event}
Recall that $L = \log\bigl( 10HS^2AK^4 \bigr)$.
We define the high probability good event
\begin{equation}\label{eq:definition_econc}
    \econc
    \coloneq
    \left\{
    \begin{aligned}
        &\abs*{\hat P_{h,s,a,n}(s')-P_{h,s,a}(s')}
        \leq\sqrt{\frac{2P_{h,s,a}(s')L}{n}}+\frac Ln
        \\
        &\hspace{1em}\forall(h,s,a,s',n)\in[H]\times\St\times\A\times\St\times[K]
    \end{aligned}
    \right\}.
\end{equation}
The following lemma states that $\econc$ holds with high probability and provides three useful inequalities which hold under $\econc$.
The proof follows from Bernstein's inequality and a union bound over all $(h,s,a,s',n)$.
We defer the complete proof to Appendix~\ref{sec:econc_proof}.
\begin{lemma} \label{lem:econc}
    The event $\econc$ defined in Eq.~\eqref{eq:definition_econc} holds with probability at least $1-1/K^3$.
    Moreover, under $\econc$, the following hold simultaneously for every $(h,s,a)\in [H]\times\St\times\A$, every $n\in[K]$, and every (potentially data-dependent) $V\in[0,H]^S$:
    \begin{enumerate}
        \item $\var(P_{h,s,a}, V) \leq 2 \thin \var\bigl(\hat{P}_{h,s,a,n},V\bigr) + \frac{8H^2SL}{n}$,
        \item $\var\bigl(\hat{P}_{h,s,a,n}, V\bigr) \leq 2\thin \var(P_{h,s,a},V) + \frac{8H^2SL}{n}$,
        \item $\hat{P}_{h,s,a,n}V \leq \left(1 + \frac{1}{16H}\right) P_{h,s,a}V + \frac{9H^2SL}{n}$.
    \end{enumerate}
\end{lemma}

\subsection{Bounding \texorpdfstring{$\topt$}{Tminus}: Cost of Under-Optimism}

In this section, we bound the expectation of the cost of under-optimism, that is $\E[\topt]$.

\begin{lemma}[Expected amount of under-optimism] \label{lem:topt_bound}
    The expected cost of under-optimism is bounded by
    \begin{equation}
        \E[\topt] \leq O\parens*{\sqrt{H^3SAK}\bigr)} \med .
    \end{equation}
\end{lemma}

\subsubsection{Stability Properties of Our Bonus Function}

We start by providing various stability results on the bonus function.
The goal of the present section is to prove Lemma~\ref{lem:sketch_bonus_stability}, restated in the form of Corollary~\ref{cor:stability}.
The following lemma controls how much the bonus changes when the value function increases.
We defer the proof to Section~\ref{sec:ordered_stability_proof}.

\begin{lemma}[Stability of the Bonus Function] \label{lem:ordered_stability}
    Let $p\in\Delta(\St)$ and $n\geq 1$ be arbitrary.
    For any $0 \leq V \leq V' \leq H\one$, 
    \[
    \bigl|b(p,n,V') - b(p,n,V)\bigr| 
    \leq 
    \frac{1}{16H} p(V'-V)
    \med .    
    \]
\end{lemma}

As an immediate consequence, the corresponding Bellman update has the following monotonicity property.

\begin{corollary}[Monotonicity of the Bellman Update] \label{cor:monotonicity}
    Let $p\in\Delta(\St)$ be a probability distribution and $n\geq 1$ be arbitrary.
    For any vectors $0 \leq V \leq V' \leq H\one$, and any $c\in(0,16)$, we have
    \[
    pV' + c\cdot b(p,n,V') 
    \geq 
    pV + c\cdot b(p,n,V)
    \med .
    \]
\end{corollary}

\begin{proof}
    For $0 \leq V \leq V' \leq H\one$ and $c\in(0,16)$, 
    \begin{align*}
    pV + c\cdot b(p,n,V) 
    &= pV' + c\cdot b(p,n,V') + p(V - V') + c\cdot b(p,n,V) - c\cdot b(p,n,V') \\
    & \overset{\eqnum{1}}\leq pV' + c\cdot b(p,n,V') + \Bigl(1 - \frac{c}{16H}\Bigr)p(V - V')  \\
    & \overset{\eqnum{2}}\leq pV' + c\cdot b(p,n,V')
    \end{align*}
    where 
    \texteqnum{1} follows by Lemma~\ref{lem:ordered_stability};
    and 
    \texteqnum{2} uses that $V - V' \le 0$ and that $1 - c / (16 H) \ge 0$.
\end{proof}


Of course, we need something that holds when $V$ and $V'$ are not necessarily ordered.
It is not hard to generalize the previous lemma to this case by considering $V$ and $V + (V' - V)_+$, see below.

\begin{corollary}[Stability of the Bellman Update, Complete Version of Lemma~\ref{lem:sketch_bonus_stability}] \label{cor:stability}
    Let $p\in\Delta(\St)$ and $n\geq 1$ be arbitrary. For any $V,V'\in[0,H]^\St$, and any $c\in(0,16)$, we have
    \[
    pV + c\cdot b(p,n,V) 
    \geq 
    pV' + c\cdot b(p,n,V') - \left(1+ \frac{c}{16H}\right) p(V'-V)_+.
    \]
\end{corollary}

\begin{proof}
The idea is to apply previous results with $V$ and $V + \pospart{V' - V}$.
We have
\begin{align*}
    pV + c\cdot b(p,n,V)
    & \overset{\eqnum{1}}\geq pV + c\cdot b\left( p,n,V + (V' - V)_+ \right) - \frac{c}{16H}p(V' - V)_+ \\
    &= p\left( V + (V' - V)_+ \right) + c\cdot b\left( p,n,V + (V' - V)_+ \right) - \left(1 + \frac{c}{16H} \right) p(V' - V)_+ \\
    & \overset{\eqnum{3}}\geq pV' + c\cdot b(p,n,V') - \left( 1 + \frac{c}{16H} \right) p(V' - V)_+
\end{align*}
where 
\texteqnum{1} follows by Lemma~\ref{lem:ordered_stability};
and 
\texteqnum{3} holds by $V + (V' - V)_+ \geq V'$ and Corollary~\ref{cor:monotonicity}.
\end{proof}

\subsubsection{Proof of Lemma~\ref{lem:topt_bound}}

We now move to the proof of Lemma~\ref{lem:topt_bound}.
Let $\mathcal{E}$ be the good event of Lemma~\ref{lem:econc}.

Recall that $\bbar = b/2$.
Writing $c_H = 1 + 1/(32H)$ and using $b \geq b_2$, Corollary~\ref{cor:stability} with $c = 1/2$ gives
\begin{align*}
    \vstar_h(s) - \vkh(s) 
    &\leq P_{h,s,\pistar_h(s)}\vstar_{h+1} - P^k_{h,s,\pistar_h(s)}\vkhp - b\left(P^k_{h,s,\pistar_h(s)},\nkh(s,\pistar_h(s)),\vkhp\right) \\
    &\leq P_{h,s,\pistar_h(s)}\vstar_{h+1} - P^k_{h,s,\pistar_h(s)}\vkhp - \bbar\left(P^k_{h,s,\pistar_h(s)},\nkh(s,\pistar_h(s)),\vkhp\right) - \frac{\beta H^2SL}{2\nkh(s,\pistar_h(s))}\\
    &\leq P_{h,s,\pistar_h(s)}\vstar_{h+1} - P^k_{h,s,\pistar_h(s)}\vstar_{h+1} - \bbar\left(P^k_{h,s,\pistar_h(s)},\nkh(s,\pistar_h(s)),\vstar_{h+1} \right)  \\
    &\qquad\qquad\qquad\qquad\qquad\qquad\qquad\qquad\qquad + 
    \underbrace{
        c_H P^k_{h,s,\pistar_h(s)}(\vstar_{h+1}-\vkhp)_+ - \frac{\beta H^2SL}{2\nkh(s,\pistar_h(s))}
    }_{\mathrm{Term~(I)}} \med .
\end{align*}
At this point the next-step error is weighted by the empirical transition $P^k_{h,s,\pistar_h(s)}$. 
Unrolling the induction on $h$ now would produce data-dependent visitation probabilities. 
Instead, we use the correction term split off from the bonus to replace $P^k_{h,s,\pistar_h(s)}$ by the true transition $P_{h,s,\pistar_h(s)}$, and only then we proceed with the induction.
Assume that $\smash{\Nkh(s,\pistar_h(s)) \geq 1}$.
Then, on $\econc$, we have 
\begin{align*}
\mathrm{Term~(I)}
& \overset{\eqnum{1}}\leq c_H\left(1 + \frac{1}{16H}\right) P_{h,s,\pistar_h(s)}(\vstar_{h+1} - \vkhp)_+ + \left(9c_H - \frac{\beta}{2} \right)\frac{H^2SL}{\nkh(s,\pistar_h(s))} \\
& \overset{\eqnum{2}}\leq \left(1 + \frac{1}{4H}\right) P_{h,s,\pistar_h(s)}(\vstar_{h+1} - \vkhp)_+
\end{align*}
where
\texteqnum{1} follows by Part 3 of Lemma~\ref{lem:econc};
and
\texteqnum{2} holds since $9 c_H \le \frac 12 \beta$ by choice of $\beta$.
In the case that $\Nkh(s,\pistar_h(s)) = 0$, then $(\vstar_{h+1} - \vkhp)_+ = 0$ and the above inequality holds trivially.
Combining with the previous display inequality and taking the positive parts of both sides, we obtain
\begin{align*}\label{eq:true-recursion}
\bigl(\vstar_h(s) - \vkh(s)\bigr)_+
&\leq 
\left((P_{h,s,\pistar_h(s)} - P^k_{h,s,\pistar_h(s)}) \vstar_{h+1} - \bbar\left(P^k_{h,s,\pistar_h(s)}, \nkh(s,\pistar_h(s)), \vstar_{h+1}\right)\right)_+  \\
&\qquad\qquad\qquad\qquad\qquad\qquad\qquad\qquad\qquad\qquad + \left(1 + \frac{1}{4H}\right) P_{h,s,\pistar_h(s)}(\vstar_{h+1} - \vkhp)_+
\med .
\end{align*}
Observing the recursive nature of this inequality, we can iterate backwards and use the fact that $(1+\frac{1}{4H})^H \leq 2$ to obtain that, on $\econc$, for every fixed initial state $s_1$,
\begin{equation} \label{eq:unrolled}
\left(\vstar_1(s_1) - \vko(s_1)\right)_+
\leq
2 \sum_{h=1}^H \sum_{s\in\St} \P^{\pistar}_{s_1}(s_h=s) \left((P_{h,s,\pistar_h(s)} - P^k_{h,s,\pistar_h(s)}) \vstar_{h+1} - \bbar\left(P^k_{h,s,\pistar_h(s)}, \nkh(s,\pistar_h(s)), \vstar_{h+1} \right)\right)_+,
\end{equation}
where $\P^{\pistar}_{s_1}(s_h=s)$ denotes the probability of visiting $s$ at step $h$ in the true MDP, starting from $s_1$ and following the policy $\pistar$.

Since the weights in the above display are deterministic for a fixed $s_1$, it suffices to bound the expected shortfall for each fixed $(k,h,s,a)$ \emph{instead of requiring all shortfalls to vanish simultaneously}.
That is, no union bound is necessary.

The following lemma gives us the desired bound on the expected shortfall for a fixed $(k,h,s,a)$.
The proof, which we give in Section~\ref{sec:expected_shortfall_proof}, follows from the time-uniform concentration inequality Theorem~\ref{thm:moss_conc} by integrating its tail bound.

\begin{lemma}[Expected Shortfall] \label{lem:row_shortfall}
    For every $k\in[K]$, $h\in[H]$, $s\in\St$, and $a\in\A$,
    \[
    \E\left[ \ind(\econc) \left( (P_{h,s,a} - P^k_{h,s,a})\vstar_{h+1} - \bbar(P^k_{h,s,a},\nkh(s,a), \vstar_{h+1}) \right)_+ \right]
    \lesssim
    \frac{\varp(P_{h,s,a},\vstar_{h+1})}{\sqrt{B}}.
    \]
\end{lemma}

Now, fix $k$ and a deterministic initial state $s_1$.
Applying Lemma~\ref{lem:row_shortfall} to \eqref{eq:unrolled} yields
\begin{equation*}
\E\bigl[\ind(\econc)(V_1^\star(s_1)-V_1^k(s_1))_+\bigr]
\lesssim
\frac{1}{\sqrt{B}} \sum_{h=1}^H \sum_{s\in\St} \P_{s_1}^{\pistar}(s_h=s) \varp\bigl( P_{h,s,\pistar_h(s)}, \vstar_{h+1}\bigr).
\end{equation*}
Then using Lemma~\ref{lem:econc} to bound $\P(\econc^c) \leq 1/K^3$,
\begin{equation*}
\E\bigl[(V_1^\star(s_1)-V_1^k(s_1))_+\bigr]
\lesssim
\frac{1}{\sqrt{B}} \sum_{h=1}^H \sum_{s\in\St} \P_{s_1}^{\pistar}(s_h=s) \varp\bigl( P_{h,s,\pistar_h(s)}, \vstar_{h+1}\bigr) + \frac{H}{K^3}.
\end{equation*}

This is for a fixed state $s_1$ instead of the observed $\sko$ which is randomly drawn from $\mu$.
Let $\P^{\pistar}_\mu(s_h=s)$ denote the probability of visiting $s$ at step $h$ in the true MDP, starting from a state randomly drawn from $\mu$ then following $\pistar$.
Denote by $\smash{\E^{\pistar}_\mu [-]}$ the associated expectation operator.
Because $\vko$ is computed from the history before $\sko\sim \mu$ is drawn, it is independent of $\sko$. 
Subsequently, since $\smash{\P^{\pistar}_\mu(s_h=s) = \sum_{s_1} \mu(s_1) \P^{\pistar}_{s_1}(s_h=s)}$,
\begin{align}\label{eq:random_start}
\E\bigl[(\vstar_1(\sko)-\vko(\sko))_+\bigr]
&= \sum_{s_1\in\St} \mu(s_1)\E\left[(\vstar_1(s_1)-\vko(s_1))_+\right] \nonumber \\
&\lesssim \frac{1}{\sqrt{B}} \sum_{h=1}^H \sum_{s\in\St} \P_{\mu}^{\pistar}(s_h=s) \varp\bigl(P_{h,s,\pistar_h(s)},\vstar_{h+1}\bigr) + \frac{H}{K^3}.
\end{align}
To bound the weighted sum of variances, observe that, along an episode generated by $\pistar$, the optimal Bellman equation and telescoping give
\begin{align*}
    &\sum_{h=1}^H \sum_{s\in\St} \P^{\pistar}_\mu(s_h=s) \Var(P_{h,s,\pistar_h(s)}, \vstar_{h+1}) \\
    &\wide = \E^{\pistar}_\mu \sum_{h=1}^H \left[ P_{h,s_h,\pistar_h(s_h)} (\vstar_{h+1})^2 - \bigl(\vstar_h(s_h) - r_h(s_h,\pistar_h(s_h)) \bigr)^2 \right] \\
    &\wide = -\E_{s_1\sim\mu}\left[(\vstar_1(s_1))^2\right] + \E^{\pistar}_\mu \left[\sum_{h=1}^H \left[ 2 r_h(s_h,\pistar_h(s_h)) \vstar_h(s_h) -  r_h(s_h,\pistar_h(s_h))^2 \right]\right] \\
    &\wide \leq 2H\,\E^{\pistar}_\mu \left[\sum_{h=1}^H r_h(s_h,\pistar_h(s_h))\right] \\
    &\wide \leq 2H^2.
\end{align*}
Since $\varp(p,V) \leq \var(p,V)+1$ and
$\sum_s\P^{\pistar}_\mu(s_h=s)=1$, it follows that
\[
    \sum_{h=1}^H\sum_{s\in\mathcal S}\P^{\pistar}_\mu(s_h=s) \varp(P_{h,s,\pi_h^\star(s)},V_{h+1}^\star)
    \leq
    2H^2+H
    \le
    3H^2.
\] 
Substituting back into \eqref{eq:random_start} and plugging in $B=HK/(SA)$ gives us that
\[
    \E\left[(\vstar_1(\sko) - \vko(\sko))_+\right]
    \lesssim
    \frac{H^2}{\sqrt{B}} + \frac{H}{K^3}
    \lesssim
    \sqrt{H^3SA/K}.
\]
By summing over $k$, we have completed the proof of Lemma~\ref{lem:topt_bound}.

\subsection{Bounding \texorpdfstring{$\tmodel$}{Tplus}: Expected Amount of Over-optimism}
\label{sec:appendix_amount_over_optimism}

In this section, we prove the following result.
\begin{lemma}[Expected amount of over-optimism]
\label{lem:bound_tmodel}
    The amount of over-optimism is bounded by
    \begin{equation}
        \E[\tmodel] = O \parens*{\sqrt{H^3SAK} + H^3S^2AL^2}
        \med .
    \end{equation}
\end{lemma}
To prove Lemma~\ref{lem:bound_tmodel}, we further decompose $\tmodel$ into three terms:
\[
\tmodel
\leq
\underbrace{\sum_{k,h} \left( (P^k_{h,\skh,\akh} - P_{h,\skh,\akh}) \vstar_{h+1} \right)_+}_{\eqcolon \tmodelstar}
+
\underbrace{\sum_{k,h} \left| (P^k_{h,\skh,\akh} - P_{h,\skh,\akh})(\vkhp - \vstar_{h+1}) \right|}_{\eqcolon \tmodeldiff}
+
\underbrace{\sum_{k,h} \bkh(\skh,\akh)}_{\eqcolon \tmodelbonus}.
\]
The term $\tmodelstar$ measures transition estimation error applied to the optimal value function, while $\tmodeldiff$ measures transition error applied to the difference $\vkhp-\vstar_{h+1}$.
The remaining term $\tmodelbonus$ is the sum of bonuses over the learner's trajectory.
We will additionally make use of the following two variance sums:
\[
\vardiff \coloneq \sum_{k,h} \var\left( P_{h,\skh,\akh}, \vkhp - \vstar_{h+1} \right)
, \qquad
\varstar \coloneq \sum_{k,h} \varp\left( P_{h,\skh,\akh}, \vstar_{h+1} \right).
\]

\subsubsection{A Few Useful Technical Results}

The following lemma, which we prove in Appendix~\ref{sec:var_sum_proof}, gives a bound on $\E[\varstar]$.
We deal with $\vardiff$ later.

\begin{lemma}[Sum of variances]
\label{lem:var_sum}
    For the total visit counts $N_h(s,a)$ defined above, we have
    \begin{equation*}
        \E \brackets*{
            \sum_{h,s,a}
            \var(P_{h,s,a}, \vstar_{h+1})
            N_h(s,a) 
        }
        \le 4 H^2K.
    \end{equation*}
    Consequently, the sum of floored variances satisfies
    \[
        \E[\varstar] \leq 5H^2K.
    \]
\end{lemma}

We also frequently use the following deterministic sum.
The proof is given in Appendix~\ref{sec:counts_sum_proof}.

\begin{lemma} \label{lem:counts_sum}
    $\sum_{k,h} \frac{1}{\nkh(\skh,\akh)} \leq HSAL$.
\end{lemma}

\subsubsection{Analysis of \texorpdfstring{$\tmodelstar$}{T1+}}

We can now proceed to analyzing the first term $\tmodelstar$.
We show the following bound.

\begin{lemma}[Bound on $\tmodelstar$]
\label{lem:tmodel_star_bound}
    We have $\E[\tmodelstar] \leq O\left( \sqrt{H^3SAK} + H^2SA \right)$.
\end{lemma}

To do so, we prove a general concentration inequality -- which may be of independent interest -- involving sums of deviations.
It is related to, but different than the time-uniform Bernstein-type concentration result, which considers a maximum over deviations.
The complete proof is in Appendix~\ref{sec:sum_slack_conc_proof}.

\begin{proposition}[Proposition~\ref{prop:sketch_sum_slack}, Concentration on sums with slackness] \label{prop:sum_slack_conc}
Let $X_1,X_2,\dots$ be i.i.d.\ random variables such that
\[
\E[X_i]=\mu,
\qquad
\var(X_i)=\sigma^2,
\qquad
|X_i-\mu|\leq c \quad\text{a.s.}
\]
For $n\geq 1$, write $\hat{\mu}_n \coloneq \frac{1}{n}\sum_{i=1}^n X_i$.
There is a universal constant $C>0$ such that, for every slackness $\Delta > 0$,
\begin{equation*}
    \sum_{n=1}^\infty \E\left[ \left( \hat{\mu}_n - \mu - \Delta\right)_+ \right]
    \leq
    C\left( \frac{\sigma^2}{\Delta} + c \right).
\end{equation*}
\end{proposition}
By tuning the slackness $\Delta$, we obtain the following useful corollary.
\begin{corollary}[Corollary~\ref{cor:sketch_sum}, Concentration on sums] \label{cor:sum_conc}
    Under the setting of Proposition~\ref{prop:sum_slack_conc}, there is a universal constant $C'>0$ such that, for any nonnegative integer-valued random variable $N$ with $\E[N] < \infty$,
    \[
    \E\left[ \sum_{n=1}^N \left( \hat{\mu}_n - \mu \right)_+ \right]
    \leq
    C'\left( \sqrt{\sigma^2 \E[N]} + c\right).
    \]
\end{corollary}
\begin{proof}
The cases $\E[N]=0$ and $\sigma^2=0$ are trivial, so assume $\E[N]>0$ and $\sigma^2>0$.
By Proposition~\ref{prop:sum_slack_conc}, we have for any $\Delta>0$,
\[
\E\left[ \sum_{n=1}^N \left(\hat{\mu}_n - \mu\right)_+ \right]
\leq
\Delta \E[N] + C\left(\frac{\sigma^2}{\Delta} + c \right).
\]
Then taking $\Delta = \sqrt{\sigma^2/\E[N]}$ and $C'=C+1$ yields the desired result.
\end{proof}

\begin{proof}[Proof of Lemma~\ref{lem:tmodel_star_bound}]
For each fixed $(h,s,a)$, apply Corollary~\ref{cor:sum_conc} to the i.i.d.\ random variables $\vstar_{h+1}(X_{h,s,a,i})$, whose mean is $P_{h,s,a}\vstar_{h+1}$, variance is $\var(P_{h,s,a},\vstar_{h+1})$, and centered absolute value is at most $H$.
Since $N_h(s,a)\leq K$, the corollary gives
\[
    \E\left[\sum_{n=1}^{N_h(s,a)}
        \left((\hat P_{h,s,a,n}-P_{h,s,a})\vstar_{h+1}\right)_+\right]
    \lesssim
    \sqrt{\var(P_{h,s,a},\vstar_{h+1})\E[N_h(s,a)]}+H.
\]
It follows that
\begin{align*}
    \E[\tmodelstar]
    &= \E\left[\sum_{k,h} \left( (P^k_{h,\skh,\akh} - P_{h,\skh,\akh}) \vstar_{h+1} \right)_+\right] \\
    &\overset{\eqnum{1}}\leq H^2SA + \sum_{h,s,a} \E\left[\sum_{n=1}^{N_h(s,a)} \left( \hat{P}_{h,s,a,n}\vstar_{h+1} - P_{h,s,a}\vstar_{h+1}  \right)_+\right] \\
    &\lesssim H^2SA + \sum_{h,s,a} \sqrt{\var(P_{h,s,a},\vstar_{h+1})\E[N_h(s,a)]} \\
    &\overset{\eqnum{2}}\leq H^2SA + \sqrt{HSA \sum_{h,s,a} \var(P_{h,s,a},\vstar_{h+1})\E[N_h(s,a)]} \\
    &\overset{\eqnum{3}}\lesssim H^2SA + \sqrt{H^3SAK}.
\end{align*}
Here, \texteqnum{1} accounts for the time steps where the count is zero; \texteqnum{2} is by Cauchy-Schwarz; and \texteqnum{3} is by Lemma~\ref{lem:var_sum}.
\end{proof}

\subsubsection{Analysis of \texorpdfstring{$\tmodeldiff$}{T2+}}

Next, we bound the second term $\tmodeldiff$.
The analysis of this term is relatively standard.

\begin{lemma}[Bound on $\tmodeldiff$] \label{lem:tmodel_diff_bound}
    We have
    $\E[\tmodeldiff] \leq O\left( \sqrt{HS^2AL^2 \E[\vardiff]} + H^2S^2AL^2 \right)$.
\end{lemma}

\begin{proof}[Proof of Lemma~\ref{lem:tmodel_diff_bound}]
Let $\econc$ be the good event of Lemma~\ref{lem:econc}.
Under $\econc$, for any $h\in[H],s\in\St,a\in\A,n\in [K]$ and (possibly data-dependent) $V\in[-H,H]^\St$, we have
\begin{align*}
    \left|\bigl(\hat P_{h,s,a,n}-P_{h,s,a}\bigr)V\right|
    &\overset{\eqnum{1}}=\left|\bigl(\hat P_{h,s,a,n}-P_{h,s,a}\bigr)(V - P_{h,s,a}V)\right| \\
    &\overset{\eqnum{2}}\leq \sum_{s'} \left|\hat P_{h,s,a,n}(s')-P_{h,s,a}(s')\right||V(s') - P_{h,s,a}V| \\
    &\overset{\eqnum{3}}\leq \sum_{s'}\sqrt{\frac{2P_{h,s,a}(s')L}{n}}|V(s')-P_{h,s,a}V| + \frac{L}{n}\sum_{s'}|V(s')-P_{h,s,a}V| \\
    &\overset{\eqnum{4}}\leq \sqrt{\frac{2SL\var(P_{h,s,a},V)}n} + \frac{2HSL}{n}.
\end{align*}
\texteqnum{1} holds because $(\hat{P}_{h,s,a,n} - P_{h,s,a})c = 0$ for any real number $c$; \texteqnum{2} is by the triangle inequality; \texteqnum{3} uses $\econc$; and \texteqnum{4} follows from Cauchy-Schwarz and $|V(s') - P_{h,s,a}V| \leq 2H$.
Hence, we have on $\econc$
\begin{align*}
    \tmodeldiff
    &=\sum_{k,h} \left| (P^k_{h,\skh,\akh} - P_{h,\skh,\akh})(\vkhp - \vstar_{h+1}) \right| \\
    &\leq 2H^2SA + \sum_{k,h \colon \Nkh(\skh,\akh) \geq 1} \left| (P^k_{h,\skh,\akh} - P_{h,\skh,\akh})(\vkhp - \vstar_{h+1}) \right| \\
    &= 2H^2SA + \sum_{k,h \colon \Nkh(\skh,\akh) \geq 1} \left| \bigl(\hat{P}_{h,\skh,\akh,\nkh(\skh,\akh)} - P_{h,\skh,\akh}\bigr)(\vkhp - \vstar_{h+1}) \right| \\
    &\leq 2H^2SA + \sum_{k,h \colon \Nkh(\skh,\akh) \geq 1} \left( \sqrt{\frac{2SL\var(P_{h,\skh,\akh},\vkhp-\vstar_{h+1})}{\Nkh(\skh,\akh)}} + \frac{2HSL}{\Nkh(\skh,\akh)} \right) \\
    &\leq 2H^2SA + \sum_{k,h} \sqrt{\frac{2SL\var(P_{h,\skh,\akh},\vkhp-\vstar_{h+1})}{\nkh(\skh,\akh)}} + \sum_{k,h} \frac{2HSL}{\nkh(\skh,\akh)} \\
    &\overset{\eqnum{5}}\leq 2H^2SA + \sqrt{\sum_{k,h} 2SL\var(P_{h,\skh,\akh},\vkhp-\vstar_{h+1})} \sqrt{\sum_{k,h} \frac{1}{\nkh(\skh,\akh)}} + \sum_{k,h} \frac{2HSL}{\nkh(\skh,\akh)} \\
    &\overset{\eqnum{6}}\leq \sqrt{2HS^2AL^2\vardiff} + 4H^2S^2AL^2,
\end{align*}
where \texteqnum{5} is due to Cauchy-Schwarz, and \texteqnum{6} is due to Lemma~\ref{lem:counts_sum} (which bounds $\sum_{k,h} 1/\nkh(\skh,\akh) \leq HSAL$) and the definition of $\vardiff$.

When $\econc$ does not hold, we have the trivial bound $\tmodeldiff \leq 2H^2K$.
Combining this bound with $\P(\econc^c) \leq \frac{1}{K^3}$ from Lemma~\ref{lem:econc}, we conclude
\begin{align*}
    \E[\tmodeldiff]
    &= \E[\ind(\econc)\tmodeldiff] + \E[\ind(\econc^c)\tmodeldiff] \\
    &\leq \sqrt{2HS^2AL^2} \E\left[\ind(\econc)\sqrt{\vardiff}\right] +4H^2S^2AL^2+2H^2K\P(\econc^c) \\
    &\overset{\eqnum{7}}\leq \sqrt{2HS^2AL^2\,\E\left[\vardiff\right]} + 4H^2S^2AL^2 + \frac{2H^2}{K^2} \\
    &\lesssim \sqrt{HS^2AL^2\E\left[\vardiff\right]} + H^2S^2AL^2.
\end{align*}
\texteqnum{7} uses $\E[\ind(\econc)\sqrt{\vardiff}] \leq \E[\sqrt{\vardiff}]\leq\sqrt{\E[\vardiff]}$ by Jensen's inequality.
\end{proof}

\subsubsection{Analysis of \texorpdfstring{$\tmodelbonus$}{T3+}}

The bound on the third term $\tmodelbonus$ requires the following key lemma.
We will use it to show that our MOSS-style bonus can be summed without introducing log factors.

\begin{lemma}[Variance-Normalized MOSS Sum] \label{lem:deterministic_moss_sum}
    For any fixed $q > 0$, $B>0$, and integer $N\geq 0$, we have
    \[
    \sum_{n=1}^N \sqrt{\frac{q\logpp \frac{B}{nq}}{n}}
    \leq
    6\sqrt{B} + \frac{qN}{\sqrt{B}}.
    \]
\end{lemma}

\begin{proof}
Write $m = \lfloor B/q \rfloor$.
Interpret empty sums below as zero.
We upper bound the sum from $1$ to $N$ by the sum from $1$ to $m$ plus the sum from $m+1$ to $N$.
Now, since $x\mapsto \sqrt{\frac{q\logpp \frac{B}{xq}}{x}}$ is nonincreasing on $(0,\infty)$ and $m\leq B/q$, we can bound
\begin{align*}
    \sum_{n=1}^m \sqrt{\frac{q\logpp \frac{B}{nq}}{n}}
    \leq \int_0^{B/q} \sqrt{\frac{q\logpp \frac{B}{xq}}{x}} \, dx 
    &= \sqrt{B} \int_0^1 \sqrt{\frac{\logpp\frac{1}{u}}{u}} \, du \\
    &\leq \sqrt{B} \int_0^1 \sqrt{\frac{1+\log\frac{1}{u}}{u}} \, du \\
    &= \sqrt{B} \int_0^\infty e^{-t/2}\sqrt{1+t} \, dt
    \leq 6\sqrt{B}.
\end{align*}
Above we used the substitutions $u = xq/B$ and $t = -\log u$.
For every remaining index $m < n \leq N$, we have $n > B/q$ and hence $\logpp\frac{B}{nq} = 1$, so
\[
    \sum_{n=m+1}^N \sqrt{\frac{q\logpp \frac{B}{nq}}{n}}
    = \sum_{n=m+1}^N \sqrt{\frac{q}{n}} 
    \leq N \sqrt{\frac{q}{B/q}} 
    = \frac{qN}{\sqrt{B}}.
\]
Combining the two sums yields the desired bound.
\end{proof}

With this lemma we can obtain the following bound on $\tmodelbonus$.
The main ideas here are as follows.
We use $\econc$ so that the empirical variances are approximately the same as the true variances up to an additive lower order term.
Then, we split terms involving $\vkhp$ into terms involving $\vstar$ and terms involving $\vkhp - \vstar$.
The former we bound by $O(\sqrt{H^3SAK} + H^3S^2AL^2)$ by direct computation on top of Lemma~\ref{lem:deterministic_moss_sum}, and the latter we bound by a term involving $\vardiff$.
For the complete proof, see Appendix~\ref{sec:tmodel_bonus_bound}.

\begin{lemma} \label{lem:tmodel_bonus_bound}
    $\E[\tmodelbonus] \leq O\left( \sqrt{H^3SAK} + \sqrt{HSAL^2\E[\vardiff]} + H^3S^2AL^2 \right)$.
\end{lemma}

\subsubsection{Conclusion: Bounding \texorpdfstring{$\tmodel$}{T+} and Proving Lemma~\ref{lem:bound_tmodel}}

Combining Lemma~\ref{lem:tmodel_star_bound} on $\tmodelstar$, Lemma~\ref{lem:tmodel_diff_bound} on $\tmodeldiff$, and Lemma~\ref{lem:tmodel_bonus_bound} on $\tmodelbonus$, we obtain
\begin{equation} \label{eq:tmodel_with_vardiff}
\E[\tmodel]
\leq
O\left( \sqrt{H^3SAK} + H^3S^2AL^2 + \sqrt{HS^2AL^2 \E[\vardiff]} \right).
\end{equation}
The following lemma, which we prove in Appendix~\ref{sec:vardiff_bound_proof}, bounds $\vardiff$.
\begin{lemma} \label{lem:vardiff_bound}
    We have 
    $\E[\vardiff] \leq 4H \E[\tmodel] + 2H \E[\topt]$.
\end{lemma}

Combining Lemma~\ref{lem:vardiff_bound} with \eqref{eq:tmodel_with_vardiff} and the bound on $\topt$ (Lemma~\ref{lem:topt_bound}) implies
\[
\E[\vardiff] \leq O\left(H\sqrt{H^3SAK} + H^4S^2AL^2 + \sqrt{H^3S^2AL^2 \E[\vardiff]} \right).
\]
Thus, by $x\leq a + \sqrt{bx} \implies x\leq 2a+b$,
\[
\E[\vardiff] \leq O\left(H\sqrt{H^3SAK} + H^4S^2AL^2\right).
\]
Then, plugging back into \eqref{eq:tmodel_with_vardiff} gives us
\begin{align} \label{eq:tplus_bound}
    \E[\tmodel]
    &\leq O\left( \sqrt{H^3SAK} + H^3S^2AL^2 + \sqrt{H^2S^2AL^2 \sqrt{H^3SAK}} \right) \nonumber \\
    &\leq O\left( \sqrt{H^3SAK} + H^3S^2AL^2 \right),
\end{align}
where the second inequality holds by $\sqrt{xy}\leq x + y$ with $x=H^2S^2AL^2$ and $y=\sqrt{H^3SAK}$.

\subsection{Conclusion: Bounding the Expected Regret and Proving Theorem~\ref{thm:minimax_regret}}

We conclude by the regret decomposition (Lemma~\ref{lem:regret_decomposition}), the bound on $\E[\topt]$ (Lemma~\ref{lem:topt_bound}), and the bound on $\E[\tmodel]$ (Eq.~\eqref{eq:tplus_bound}) that
\[
\E[\reg(K)]
\leq
\E[\topt] + \E[\tmodel]
\leq
O\left( \sqrt{H^3SAK} + H^3S^2AL^2 \right).
\]
If $K\leq HSA$, then $\E[\reg(K)]\leq HK\leq\sqrt{H^3SAK}$.
Otherwise, $K>HSA$ implies $K\geq2$ and $HS^2A\leq(HSA)^2<K^2$, so $L=\log(10HS^2AK^4)\leq\log(10K^6)\lesssim\log K$.
In either case,
\[
\E[\reg(K)]
\leq
O\left( \sqrt{H^3SAK} + H^3S^2A(\log K)^2 \right).
\]

\section{Missing Proofs from Section~\ref{sec:complete_proof}}

This section contains proofs which were omitted for ease of presentation.
We also state Bernstein's inequality here for completeness.

\begin{lemma}[Bernstein's Inequality, Theorem 3 in \cite{maurer2009empiricalbernsteinboundssample}]
\label{lem:bernstein}
    Let $(X_k)_{k=1}^n$ be a sequence of i.i.d.~random variables.
    Let $\mu := \E[X_1]$ be their expected value and $v := \var(X_1)$ be their variance.
    If $X_1 - \mu \le c$ a.s. for some $c>0$, then for all $x \ge 0$,
    \begin{equation*}
        \P \parens*{
            \frac 1n \sum_{k=1}^n \parens*{X_k - \mu}
            \ge
            \sqrt{\frac{2 v x}n}
            + \frac {c x}{3 n}
        }
        \le 
        \ee^{-x} 
        \med .
    \end{equation*}
\end{lemma}

\subsection{Proof of Regret Decomposition, Lemma~\ref{lem:regret_decomposition}}
\label{sec:regret_decomposition_proof}

We decompose 
\begin{align*}
    \reg(K)
    &= \sum_{k=1}^K \left( \vstar_1(\sko) - \vpiko(\sko) \right) \\
    &\leq \sum_{k=1}^K \left( \vstar_1(\sko) - \vko(\sko) \right)_+ + \sum_{k=1}^K \left( \vko(\sko) - \vpiko(\sko) \right).
\end{align*}
The first sum is precisely $\topt$, so it remains to bound the expectation of the second sum by $\E[\tmodel]$.
Towards this end, fix arbitrary $k\in[K]$.
We have
\begin{align*}
    \E\left[ \vko(\sko) - \vpiko(\sko) \right]
    &= \E\left[\vko(\sko) - \sum_{h=1}^H r_h(\skh,\akh)\right] + \E\left[ \sum_{h=1}^H r_h(\skh,\akh) - \vpiko(\sko) \right] \\
    &= \E\left[\vko(\sko) - \sum_{h=1}^H r_h(\skh,\akh)\right].
\end{align*}
Moreover, using $\vkHp = 0$, telescoping gives us
\begin{align*}
    &\vko(\sko) - \sum_{h=1}^H r_h(\skh,\akh) \\
    &= \sum_{h=1}^H \left( \vkh(\skh) - r_h(\skh,\akh) - \vkhp(\skhp) \right) \\
    &= \sum_{h=1}^H \left( P_{h,\skh,\akh}\vkhp - \vkhp(\skhp) \right) + \sum_{h=1}^H \left(\vkh(\skh) - r_h(\skh,\akh) - P_{h,\skh,\akh}\vkhp \right).
\end{align*}
The first sum has expectation 0, and for the second sum we use
\[
\vkh(\skh) = \qkh(\skh,\akh) \leq r_h(\skh,\akh) + P^k_{h,\skh,\akh}\vkhp + \bkh(\skh,\akh).
\]
Hence,
\[
    \E\left[ \vko(\sko) - \sum_{h=1}^H r_h(\skh,\akh) \right]
    \leq
    \E \left[ \sum_{h=1}^H \left( \left( P^k_{h,\skh,\akh} - P_{h,\skh,\akh} \right)\vkhp + \bkh(\skh,\akh) \right)_+ \right].
\]
Summing over $k$ completes the argument.
\qed

\subsection{Proof of Probability of the Good Event \texorpdfstring{$\econc$}{Econc}, Lemma~\ref{lem:econc}}
\label{sec:econc_proof}

    We begin by proving that $\P (\econc) \ge 1 - 1 / K^3$. 
    We have
    \begin{align*}
        \P \parens*{(\econc)^\complement}
        & \overset{\eqnum{1}}\le
        \sum_{h, s, a, s'}
        \sum_{n=1}^K
        \P \parens*{
            \abs*{
                \hat{P}_{h, s, a, n}(s')
                - P_{h, s, a}(s')
            }
            >
            \sqrt{
                \frac{2 P_{h, s, a}(s') L}n
            }
            + \frac Ln
        }
        \\ & \overset{\eqnum{2}}\le
        2 H S^2 A K e^{-L}
        \\ & \overset{\eqnum{3}}\le
        \frac 1{K^3}
    \end{align*}
    where
    \texteqnum{1} follows by a union bound;
    \texteqnum{2} follows by bounding the summand via Bernstein's inequality (Lemma~\ref{lem:bernstein}) and bounding the number of terms by $H S^2 A K$;
    and 
    \texteqnum{3} holds by the definition of $L$.

    We continue with the proof of Assertion (\emph{1.}).
    By the AM--GM inequality, $\sqrt{2 p L / n} \le p / 2 + L / n$.
    So, on $\econc$, for every $(h, s, a, s', n) \in [H]\times\St\times\A\times\St\times[K]$, we have
    \begin{equation}
    \label{eq:lem_econc_1}
        \hat{P}_{h, s, a, n} (s')
        \le 
        \frac 32 P_{h, s, a}(s')
        + \frac{2 L}n
        \quad\text{and}\quad
        P_{h, s, a} (s')
        \le 
        2 \hat{P}_{h, s, a, n}(s')
        + \frac{4 L}n
        \med .
    \end{equation}
    From there, we prove Assertion (\emph{1.}) as such:
    \begin{align*}
        \var (P_{h, s, a}, V)
        & =
        \min_{x \in \R}
        \sum_{s'} P_{h, s, a}(s')
        \parens*{ V (s') - x }^2
        \\ & \le 
        \sum_{s'} P_{h, s, a}(s')
        \parens*{ V (s') - \hat{P}_{h, s, a, n}V }^2
        \\ & \overset{\eqnum{4}}\le
        \sum_{s'} 
        \parens*{ 2 \hat{P}_{h, s, a, n}(s') + \frac{4L}n }
        \parens*{ V (s') - \hat{P}_{h, s, a, n}V }^2
        \\ & \overset{\eqnum{5}}\le
        2 \thin \var(\hat{P}_{h, s, a, n}, V)
        + \frac{4 S H^2 L}n
    \end{align*}
    where
    in \texteqnum{4} we bound $P_{h,s,a}(s')$ via Eq.~\eqref{eq:lem_econc_1};
    and 
    in \texteqnum{5} we use that $\abs{V(s') - \hat{P}_{h,s,a,n} V} \le H$ for all $s'$, which holds true since $V \in [0, H]^S$. 

    We prove Assertion (\emph{2.}) similarly, except now we use the first bound in Eq.~\eqref{eq:lem_econc_1}:
    \begin{align*}
        \var(\hat P_{h,s,a,n},V)
        &\leq\sum_{s'}\hat P_{h,s,a,n}(s')\bigl(V(s')-P_{h,s,a}V\bigr)^2
        \\
        &\leq\sum_{s'}\left(\frac32P_{h,s,a}(s')+\frac{2L}{n}\right)
        \bigl(V(s')-P_{h,s,a}V\bigr)^2
        \\
        &\leq\frac32\var(P_{h,s,a},V)+\frac{2H^2SL}{n}
        \\
        &\leq2\thin\var(P_{h,s,a},V)+\frac{8H^2SL}{n}.
    \end{align*}

    We are left with the proof of Assertion (\emph{3.}):
    \begin{align*}
        \hat{P}_{h, s, a, n} V
        & =
        P_{h, s, a} V
        + \sum_{s'} 
        \parens*{
            \hat{P}_{h, s, a, n}(s') 
            - P_{h, s, a} (s')
        }
        V(s')
        \\ & \overset{\eqnum{6}}\le
        P_{h, s, a} V
        + \sum_{s'} 
        \parens*{
            \sqrt{\frac{2 P_{h, s, a}(s') L}n}
            + \frac Ln
        }
        V(s')
        \\ & \overset{\eqnum{7}}\le
        P_{h, s, a} V
        + \sum_{s'} 
        \parens*{
            \frac {P_{h, s, a}(s')}{16H}
            + \frac{(8 H + 1)L}n
        }
        V(s')
        \\ & \overset{\eqnum{8}}\le
        \parens*{1 + \frac1{16H}} P_{h, s, a} V
        + \frac{9 H^2 S L}n
    \end{align*}
    where 
    \texteqnum{6} follows by $V \ge 0$ and because we are on $\econc$;
    \texteqnum{7} follows from the AM--GM inequality $\sqrt{2 p L / n} \le p / (16 H) + 8 H L / n$;
    and
    \texteqnum{8} uses that $V(s') \le H$ and $H \ge 1$.
\qed

\subsection{Proof of Ordered Stability, Lemma~\ref{lem:ordered_stability}}
\label{sec:ordered_stability_proof}

Fix arbitrary $p\in\Delta(\St)$, $n\geq 1$, and
$0 \leq V \leq V' \leq H\one$.
We first observe that
\begin{align*}
    \abs*{\wtvar(p,n,V') - \wtvar(p,n,V)}
    &\overset{\eqnum{1}}{=}
    2\abs*{\varp(p,V') - \varp(p,V)} \\
    &\overset{\eqnum{2}}{\leq}
    2\abs*{\var(p,V') - \var(p,V)} \\
    &\overset{\eqnum{3}}{=}
    2\abs*{
        p\parens*{(V')^2-V^2}
        - \parens*{(pV')^2-(pV)^2}
    } \\
    &\overset{\eqnum{4}}{=}
    2\abs*{
        p\parens*{(V'-V)(V'+V)}
        - p(V'-V)p(V'+V)
    } \\
    &\overset{\eqnum{5}}{=}
    2\abs*{
        p\parens*{
            (V'-V)\parens*{V'+V-p(V'+V)\one}
        }
    } \\
    &\overset{\eqnum{6}}{\leq}
    4H p(V'-V)
    \med .
\end{align*}
Here, \texteqnum{1} unfolds the definition of $\wtvar$; 
\texteqnum{2} holds because $\abs{\max(u, 1) - \max(v, 1)} \le \abs{u - v}$ for $u, v \in \R$;
and \texteqnum{3} unfolds the definition of $\var$.
For \texteqnum{4}, the difference-of-squares identity and linearity of $p$ give
\[
    (pV')^2-(pV)^2
    =
    (pV'-pV)(pV'+pV)
    =
    p(V'-V)p(V'+V)
    \med .
\]
Then \texteqnum{5} uses that $p(V'+V)$ is a scalar and can therefore be moved
inside $p$. Finally, \texteqnum{6} follows from $V'-V\geq 0$ and
\[
    \abs*{V'+V-p(V'+V)\one} \leq 2H\one
    \med .
\]

Next, the function $\Phi(x) = x\logpp(B/(nx))$ is continuous
on $[1,\infty)$, and, wherever it is differentiable,
\[
    \frac{\mathrm{d}}{\mathrm{d}x}\parens*{
        x\logpp\parens*{\frac{B}{ nx}}
    }
    =
    \begin{cases}
        \displaystyle \log\parens*{\frac{B}{ nx}} - 1,
        & x < \dfrac{B}{ en}, \\[6pt]
        1,
        & x > \dfrac{B}{ en}.
    \end{cases}
\]
Both expressions lie in $[0,\logpp(B/ n)]$ on their respective
domains within $[1,\infty)$.

Now we argue that, for $x,x' \geq 1$ we have
\begin{equation}
    |\Phi(x') - \Phi(x)| \leq \logpp\parens*{\frac{B}{ n}} |x'-x|.\label{eq:ordered_stability_MVT_app}
\end{equation}
If $x=x'$ the bound is immediate, so otherwise assume WLOG that $x' > x$. Let $x_0 = B/( en)$. If $x_0 \not \in (x,x')$ then $\Phi$ is differentiable on the interval $(x,x')$ and the mean value theorem immediately implies~\eqref{eq:ordered_stability_MVT_app}.
If $x < x_0 < x'$ then applying the mean value theorem on each of the intervals $[x,x_0]$ and $[x_0, x']$ (noting that $\Phi$ is continuous at $x_0$) yields that
\begin{align*}
    |\Phi(x') - \Phi(x)|
    &\leq |\Phi(x') - \Phi(x_0)| + |\Phi(x_0) - \Phi(x)| \\
    &\leq |x'-x_0| + \logpp(B/n) |x_0 - x| \\
    & \leq \logpp(B/ n) |x'-x|.
\end{align*}
Thus we have shown~\eqref{eq:ordered_stability_MVT_app}.

Now, noting $b_1(p,n,V)^2 = \frac{\alpha}{n}\Phi(\wtvar(p,n,V)) $, we can apply~\eqref{eq:ordered_stability_MVT_app} to obtain that
\[
    \abs*{b_1(p,n,V')^2 - b_1(p,n,V)^2}
    \leq
    \frac{\alpha}{n}\logpp\parens*{\frac{B}{ n}}
    \abs*{\wtvar(p,n,V') - \wtvar(p,n,V)}
    \med ,
\]
where we used that $\wtvar(p,n,V)$ and $\wtvar(p,n,V')$
are both at least $2$.

Moreover, from definitions we have $\logpp(B/n) \leq 2L$ and $b_2(p,n,V') = b_2(p,n,V) = \frac{\beta H^2SL}{n} > 0$.
Thus,
\begin{align*}
    \abs*{b(p,n,V') - b(p,n,V)}
    &\overset{\eqnum{7}}{=}
    \frac{\abs*{b(p,n,V')^2 - b(p,n,V)^2}}
    {b(p,n,V') + b(p,n,V)} \\
    &\overset{\eqnum{8}}{\leq}
    \frac{\abs*{b_1(p,n,V')^2 - b_1(p,n,V)^2}}
    {2b_2(p,n,V)} \\
    &\overset{\eqnum{9}}{\leq}
    \frac{\alpha\logpp(B/n)}{2n b_2(p,n,V)}
    \abs*{\wtvar(p,n,V') - \wtvar(p,n,V)} \\
    &\overset{\eqnum{10}}{\leq}
    \frac{2\alpha H\logpp(B/n)}{n b_2(p,n,V)}
    p(V'-V) \\
    &\overset{\eqnum{11}}{=}
    \frac{2\alpha\logpp(B/n)}{\beta HSL}
    p(V'-V) \\
    &\overset{\eqnum{12}}{\leq}
    \frac{4\alpha}{\beta HS}p(V'-V) \\
    &\overset{\eqnum{13}}{\leq}
    \frac{1}{16H}p(V'-V).
\end{align*}
Here, \texteqnum{7} is the difference-of-squares identity. 
For \texteqnum{8}, observe that
\[
    b(p,n,V)^2
    =
    \max\braces*{b_1(p,n,V)^2,b_2(p,n,V)^2},
\]
and the second branch (in $b_2$) is the same for $V$ and $V'$. Thus, taking the maximum with this common term cannot increase the difference in the numerator, while both terms in the denominator are at least $b_2(p,n,V)$. 
Inequality \texteqnum{9} follows from the preceding bound on $b_1^2$, and \texteqnum{10} follows from \texteqnum{6}. 
Equality \texteqnum{11} unfolds the
definition of $b_2$, 
\texteqnum{12} uses $\logpp(B/n)\leq 2L$, and \texteqnum{13} follows by choosing $\beta\geq 64\alpha$ and using $S\geq 1$.
\qed

\subsection{Proof of Expected Shortfall, Lemma~\ref{lem:row_shortfall}}
\label{sec:expected_shortfall_proof}

Fix $k,h,s,a$.
When $\Nkh(s,a) = 0$, we have $\bbar\bigl(P^k_{h,s,a},\nkh(s,a),\vstar_{h+1}\bigr) \geq H$, and trivially the desired claim holds.
The desired claim also trivially holds if $\varp(P_{h,s,a},\vstar_{h+1})/\sqrt{B} \geq H$.
It therefore suffices to consider the case $\Nkh(s,a) \geq 1$ and $\varp(P_{h,s,a},\vstar_{h+1})/\sqrt{B} < H$.

On $\econc$, Lemma~\ref{lem:econc} gives, for every $n\in[K]$,
\[
    \wtvar(\hat{P}_{h,s,a,n},n,\vstar_{h+1}) \geq \varp(P_{h,s,a},\vstar_{h+1}).
\]
Then, since $x \mapsto x\logpp(B/(nx))$ is nondecreasing on $(0,\infty)$,
\[
    b_1(\hat{P}_{h,s,a,n},n,\vstar_{h+1})
    \geq
    \sqrt{\frac{\alpha\varp(P_{h,s,a},\vstar_{h+1})}{n}\logpp\frac{B}{n\varp(P_{h,s,a},\vstar_{h+1})}}.
\]

For
\[
    \frac{\varp(P_{h,s,a},\vstar_{h+1})}{\sqrt{B}} < x \leq H,
\]
define
\[
    \delta_x
    :=
    \frac{\bigl(\varp(P_{h,s,a},\vstar_{h+1})\bigr)^2}
    {B\bigl(\varp(P_{h,s,a},\vstar_{h+1})+Hx\bigr)}.
\]
Since $x>\varp(P_{h,s,a},\vstar_{h+1})/\sqrt{B}$ and $x\leq H$, we have
\[
    \bigl(\varp(P_{h,s,a},\vstar_{h+1})\bigr)^2
    < Bx^2
    \leq BHx,
\]
and hence $\delta_x\in(0,1)$. Moreover,
\[
    \frac{1}{n\delta_x}
    =
    \frac{B}{n\varp(P_{h,s,a},\vstar_{h+1})}
    \left(
        1+\frac{Hx}{\varp(P_{h,s,a},\vstar_{h+1})}
    \right).
\]
Thus, using $\logp(ab)\leq\logpp(a)+\log b$ for $b\geq1$,
\[
    \logp\frac{1}{n\delta_x}
    \leq
    \logpp\frac{B}{n\varp(P_{h,s,a},\vstar_{h+1})}
    +
    \log\left(
        1+\frac{Hx}{\varp(P_{h,s,a},\vstar_{h+1})}
    \right).
\]
We will use the two bounds
\[
    \log\left(
        1+\frac{Hx}{\varp(P_{h,s,a},\vstar_{h+1})}
    \right)
    \leq
    \frac{Hx}{\varp(P_{h,s,a},\vstar_{h+1})}
\]
and
\[
    \log\left(
        1+\frac{Hx}{\varp(P_{h,s,a},\vstar_{h+1})}
    \right)
    \leq
    \log(1+H^2)
    \leq 2H.
\]
Subsequently, on $\econc$, for every $n\in[K]$,
\[
\begin{aligned}
    \sqrt{
        \frac{\alpha_0\var(P_{h,s,a},\vstar_{h+1})}{n}
        \logp\frac{1}{n\delta_x}
    }
    +
    \frac{\beta_0H}{n}
    \logp\frac{1}{n\delta_x}
    &\overset{\eqnum{1}}{\leq}
    \frac{1}{4} b_1(\widehat P_{h,s,a,n},n,\vstar_{h+1})
    +
    \! \sqrt{\frac{\alpha_0Hx}{n}}
    +
    \frac{\beta_0H}{n}\logpp\frac{B}{n}
    +
    \frac{2\beta_0H^2}{n}
    \\
    &\overset{\eqnum{2}}{\leq}
    \frac{1}{4} b_1(\widehat P_{h,s,a,n},n,\vstar_{h+1})
    +
    \frac{1}{4} b_2(\widehat P_{h,s,a,n},n,\vstar_{h+1})
    +
    \frac{x}{2}
    \\
    &\overset{\eqnum{3}}{\leq}
    \bbar(\widehat P_{h,s,a,n},n,\vstar_{h+1})
    +
    \frac{x}{2}.
\end{aligned}
\]
For \texteqnum{1}, we use $\sqrt{u+v}\leq\sqrt{u}+\sqrt{v}$,
$\var(P_{h,s,a},\vstar_{h+1})\leq\varp(P_{h,s,a},\vstar_{h+1})$,
the preceding lower bound on $b_1$, and
\[
    \logpp\frac{B}{n\varp(P_{h,s,a},\vstar_{h+1})}
    \leq
    \logpp\frac Bn.
\]
For \texteqnum{2}, first note that
\[
    \sqrt{\frac{\alpha_0Hx}{n}}
    \leq
    \frac{x}{2}+\frac{\alpha_0H}{2n},
\]
while $H\geq1$, $S\geq 1$, $\logpp(B/n)\geq1$, $\logpp(B/n) \leq 2L$ and
$\beta\geq 4(\alpha_0+6\beta_0)$ imply
\[
\begin{aligned}
    \frac{\alpha_0H}{2n}
    +\frac{\beta_0H}{n}\logpp\frac Bn
    +\frac{2\beta_0H^2}{n}
    &\leq
    \frac{(\alpha_0+6\beta_0)H^2}{2n}
    \logpp\frac{B}{n}
    \\
    &\leq
    \frac{1}{4} b_2(\widehat P_{h,s,a,n},n,\vstar_{h+1}).
\end{aligned}
\]
Finally, \texteqnum{3} follows directly from
$\bbar=\max\{b_1/2,b_2/2\}$.

For fixed $(h,s,a)$, the successive transition observations can be generated in advance as an i.i.d. sequence with law $P_{h,s,a}$, and $\hat{P}_{h,s,a,n}$ is the empirical transition row after its first $n$ observations.
Their $\vstar_{h+1}$ values have mean $P_{h,s,a}\vstar_{h+1}$, variance $\var(P_{h,s,a},\vstar_{h+1})$, and absolute centered values bounded by $H$.
For positive visit counts, $P^k_{h,s,a}=\hat{P}_{h,s,a,\nkh(s,a)}$.
Therefore, on $\econc$,
\begin{align*}
    &(P_{h,s,a}-P^k_{h,s,a})\vstar_{h+1}
    - \bbar(P^k_{h,s,a},\nkh(s,a),\vstar_{h+1}) > x
\end{align*}
implies the failure event in Theorem~\ref{thm:moss_conc} with confidence $\delta_x$ and slackness $x/2$.
Since Theorem~\ref{thm:moss_conc} is uniform over $n$, it applies at the random count $\nkh(s,a)$ without conditioning on the count or on $\econc$.
We can therefore bound
\begin{align*}
    &\P\left(
        \econc,\,
        (P_{h,s,a}-P^k_{h,s,a})\vstar_{h+1}
        - \bbar(P^k_{h,s,a},\nkh(s,a),\vstar_{h+1}) > x
    \right)
    \\
    &\wide \leq
    16\delta_x\left(
        \frac{4\var(P_{h,s,a},\vstar_{h+1})}{x^2}
        + \frac{2H}{x}
    \right)
    \\
    &\wide \leq
    \frac{64\bigl(\varp(P_{h,s,a},\vstar_{h+1})\bigr)^2}{Bx^2}.
\end{align*}
The final inequality follows by substituting the definition of $\delta_x$ and using $\var(P_{h,s,a},\vstar_{h+1})\leq\varp(P_{h,s,a},\vstar_{h+1})$.

Since this holds for arbitrary $x\in(\varp(P_{h,s,a},\vstar_{h+1})/\sqrt{B},H]$, and the positive part in the statement is bounded by $H$, we compute
\begin{align*}
    &\E\left[
        \ind(\econc)
        \left(
            (P_{h,s,a}-P^k_{h,s,a})\vstar_{h+1}
            - \bbar(P^k_{h,s,a},\nkh(s,a),\vstar_{h+1})
        \right)_+
    \right]
    \\
    &\wide =
    \int_0^H
    \P\left(
        \econc,\,
        (P_{h,s,a}-P^k_{h,s,a})\vstar_{h+1}
        - \bbar(P^k_{h,s,a},\nkh(s,a),\vstar_{h+1}) > x
    \right)\,dx
    \\
    &\wide \leq
    \frac{\varp(P_{h,s,a},\vstar_{h+1})}{\sqrt{B}}
    + \frac{64\bigl(\varp(P_{h,s,a},\vstar_{h+1})\bigr)^2}{B}
    \int_{\varp(P_{h,s,a},\vstar_{h+1})/\sqrt{B}}^H \frac{1}{x^2}\,dx
    \\
    &\wide \leq
    65
    \frac{\varp(P_{h,s,a},\vstar_{h+1})}{\sqrt{B}}.
\end{align*}
This completes the proof.
\qed

\subsection{Proof of the Variance Sum Bound, Lemma~\ref{lem:var_sum}}
\label{sec:var_sum_proof}

Denote by $\Delta_{h,s,a} \coloneq \vstar_h(s) - r_h(s,a) - P_{h,s,a} \vstar_{h+1}$ the Bellman gap of the triplet $(h,s,a)$.
Note that the Bellman optimality equation implies that $\Delta_{h,s,a}\geq 0$.
For every triplet $(h,s,a)$, we have
\begin{align*}
    \var(P_{h,s,a}, \vstar_{h+1})
    & = 
    P_{h,s,a} (\vstar_{h+1})^2
    - \parens*{ P_{h,s,a} \vstar_{h+1} }^2
    \\ & =
    P_{h,s,a} (\vstar_{h+1})^2
    - (\vstar_h (s))^2
    + 2 \thin \vstar_h (s) \parens*{ r_h(s,a) + \Delta_{h,s,a} }
    - \parens*{ r_h(s,a) + \Delta_{h,s,a} }^2
    \\ & \overset{\eqnum{1}}\le
    P_{h,s,a} (\vstar_{h+1})^2
    - (\vstar_h (s))^2
    + 2 H \parens*{ r_h(s,a) + \Delta_{h,s,a} }
    \\ & \overset{\eqnum{2}}\le
    P_{h,s,a} (\vstar_{h+1})^2
    - (\vstar_h (s))^2
    + 2 H + 2 H \Delta_{h,s,a}
\end{align*}
where
\texteqnum{1} uses that $r_h(s,a) + \Delta_{h,s,a} \ge 0$ and the bound $\vstar_h \le H$;
and
\texteqnum{2} uses that $r_h(s,a) \le 1$.
Further summing over episodes $k \in [K]$ and time $h \in [H]$, we conclude the proof as such:
\begin{align*}
    & \E \brackets*{
        \sum_{h,s,a}
        \var(P_{h,s,a}, \vstar_{h+1})
        N_h(s,a)
    }
    \\ & \wide \overset{\eqnum{3}}\le
    \E \brackets*{
        \sum_{k=1}^K
        \sum_{h=1}^H
        \parens*{
            P_{h,s_h^k,a_h^k} (\vstar_{h+1})^2
            - (\vstar_h (s_h^k))^2
        }
    }
    + 2 H^2 K
    + 2 H \thin \E \brackets*{
        \sum_{k=1}^K
        \sum_{h=1}^H
        \Delta_{h,s_h^k,a_h^k}
    }
    \\ & \wide \overset{\eqnum{4}}\le
    \underbrace{
        \E \brackets*{
            \sum_{k=1}^K
            \parens*{
                P_{H,s_H^k,a_H^k} (\vstar_{H+1})^2
                - (\vstar_1 (s_1^k))^2
            }
        }
    }_{\le 0}
    + 2 H^2 K
    + 2 H \thin \underbrace{
        \E \brackets*{
            \sum_{k=1}^K
            \sum_{h=1}^H
            \Delta_{h,s_h^k,a_h^k}
        }
    }_{ = \E\brackets*{\reg(K)} \le H K}
\end{align*}
where 
\texteqnum{3} follows by invoking \texteqnum{2};
and 
\texteqnum{4} follows by simplifying the telescopic sum in the first expectation of \texteqnum{3}.
In \texteqnum{4}, the first expectation is nonpositive since $\smash{P_{H,s_H^k,a_H^k} (\vstar_{H+1})^2} - \smash{(\vstar_1 (s_1^k))^2} = - \smash{(\vstar_1 (s_1^k))^2} \le 0$; and the second expectation is equal to the expected regret, so is trivially bounded by $H K$. 

For the second part, $\varp(p,V) \leq \var(p,V) + 1$, so
\[
\E[\varstar] \leq 4H^2K + HK \leq 5H^2K.
\]
\qed

\subsection{Proof of Lemma~\ref{lem:counts_sum}}
\label{sec:counts_sum_proof}

First, observe that
\[
    \sum_{k,h} \frac{1}{\nkh(\skh,\akh)}
    \leq
    \sum_{h,s,a} \left( 1 + \sum_{n=1}^{N_h(s,a)-1} \frac{1}{n} \right)
\]
where the 1 accounts for the first visit.
Since 
\[
\sum_{n=1}^{N_h(s,a)-1} \frac{1}{n}
\leq
\sum_{n=1}^{K} \frac{1}{n}
\leq
1 + \log K,
\]
we conclude that $\sum_{k,h} \frac{1}{\nkh(\skh,\akh)} \leq HSA(2 + \log K) \leq HSAL$.
\qed

\subsection{Proof of the Concentration Result on Sums with Slackness, Proposition~\ref{prop:sum_slack_conc}} \label{sec:sum_slack_conc_proof}

Let $(X_i)_{i=1}^\infty$ be a sequence of i.i.d.~random variables and let $\E[X_i] = \mu$ and $\sigma^2 = \E[(X_i - \mu)^2]$ be their respective expectation and variance.
Further assume that $\abs{X_i - \mu} \le c$ a.s., for some $c \in \R_{\ge 0}$.
Recall that $\hat{\mu}_n \coloneq \frac 1n \sum_{i=1}^n X_i$ denotes the empirical mean after $n$ samples.
We show that, for all $\Delta > 0$,
\begin{equation}
\label{eq:sum_slack_conc_0}
    \sum_{n=1}^\infty
    \E \brackets*{
        \pospart*{
            \hat{\mu}_n - \mu - \Delta
        }
    }
    \le 
    C \parens*{
        \frac {\sigma^2}\Delta + c
    }
    \med .
\end{equation}
where $C < \infty$ is a universal constant.

Note that for $\sigma = 0$, we have $\E[ \pospart{\hat{\mu}_n - \mu - \Delta}] = 0$ for all $\Delta > 0$, so that Eq.~\ref{eq:sum_slack_conc_0} is trivial.
In the sequel, we assume that $\sigma > 0$.

Fix $\Delta > 0$.
To bound the summand $\E [ \pospart{\hat{\mu}_n - \mu - \Delta} ]$, we distinguish between two cases, depending on whether the visit count $n$ is below or above the threshold:
\begin{equation*}
    n^\star
    \coloneq
    \ceil*{
        \frac{\sigma^2}{\Delta^2}
        +
        \frac{c^2}{\sigma^2}
    }
    \med .
\end{equation*}
Low visit counts are handled differently because the tails spawned by Bernstein's inequality are inadequate and the resulting bound is not tight enough.
We therefore consider two cases.
\begin{itemize}
    \item 
        \emph{Partial sum for $n \le n^\star$.}
        For low visit counts, a simple Jensen inequality is enough:
        \begin{align}
        \nonumber
            \sum_{n=1}^{n^\star}
            \E \brackets*{
                \pospart*{
                    \hat{\mu}_n - \mu - \Delta
                }
            }
            & \overset{\eqnum{1}}\le
            \sum_{n=1}^{n^\star}
            \E \brackets[\big]{
                \abs*{
                    \hat{\mu}_n - \mu
                }
            }
            \\ & \overset{\eqnum{2}}\le
        \nonumber
            \sum_{n=1}^{n^\star}
            \sqrt{
                \var(\hat{\mu}_n)
            }
            \\ & \overset{\eqnum{3}}=
        \nonumber
            \sum_{n=1}^{n^\star}
            \sqrt{
                \frac 1n \var (X_1)
            }
            \overset{\eqnum{4}}\le
            2 \sqrt{n^\star \sigma^2}
            \overset{\eqnum{5}}= O \parens*{
                \frac{\sigma^2}\Delta
                + c
            }
        \end{align}
        where 
        \texteqnum{1} follows by linearity of $\E[-]$ together with the inequality $\pospart{x - y} \le \abs{x}$ for $y \ge 0$;
        \texteqnum{2} uses $\abs{x} = \sqrt{x^2}$ and invokes Jensen's inequality;
        \texteqnum{3} uses that $\hat{\mu}_n$ is the average of $n$ independent copies of $X_1$;
        \texteqnum{4} follows by a series--integral comparison;
        and
        \texteqnum{5} unfolds the definition of $n^\star$ and uses the sub-additivity of $x \mapsto \sqrt{x}$. 
        
    \item
        \emph{Partial sum for $n > n^\star$.}
        The bound for high visit counts is slightly more involved and relies on Bernstein's inequality.
        We have:
        \begin{align}
        \nonumber
            \sum_{n=n^\star}^\infty
            \E \brackets*{
                \pospart*{
                    \hat{\mu}_n - \mu - \Delta
                }
            }
            & \le
            \sum_{n=n^\star}^\infty
            \integral_{\Delta}^\infty
            \P \parens*{
                \hat{\mu}_n - \mu
                \ge 
                x
            }
            \med \mathrm{d}x
            \\ & \overset{\eqnum{6}}\le
        \nonumber
            \sum_{n=n^\star}^\infty
            \integral_{\Delta}^\infty
            \exp \parens*{
                - \frac{
                    n x^2
                }{
                    2 \parens*{
                        \sigma^2
                        + \frac 13 cx
                    }
                }
            }
            \med \mathrm{d}x
            \\ & \le
        \nonumber
            \sum_{n=n^\star}^\infty
            \integral_{\Delta}^\infty
            \exp \parens*{
                - n 
                \min \braces*{
                    \frac {x^2}{4 \thin \sigma^2},
                    \frac{3 x}{4c}
                }
            }
            \med \mathrm{d}x
            \\ & \overset{\eqnum{7}}\le
        \nonumber
            \sum_{n=n^\star}^\infty
            \parens*{
                \integral_{\Delta}^\infty
                \ee^{
                    - \frac {nx^2}{4 \thin \sigma^2}
                }
                \med \mathrm{d} x
                + \integral_{3 \sigma^2 / c}^\infty
                \ee^{
                    - \frac{3 n x}{4 c}
                }
                \med \mathrm{d} x
            }
            \\ & \overset{\eqnum{8}}\le
        \nonumber
            \sum_{n=n^\star}^\infty
            \parens*{
                2\Delta
                \thin \ee^{
                    - \frac {n \Delta^2}{4 \thin \sigma^2}
                }
                + \frac{4c}{3n}
                \thin \ee^{
                    - \frac{9 n \sigma^2}{4c^2}
                }
            }
            =
        \nonumber
            O \parens*{
                \frac{\sigma^2}{\Delta}
            }
            + 
            O \parens*{
                \frac{
                    c^3
                }{
                    \sigma^2
                    n^\star
                }
            }
            \overset{\eqnum{9}}=
            O \parens*{
                \frac{\sigma^2}{\Delta}
                + c
            }
        \end{align}
        where
        \texteqnum{6} invokes Bernstein's inequality (Lemma~\ref{lem:bernstein});
        \texteqnum{7} splits the integral depending on which of the quadratic and the linear terms dominate;
        and
        \texteqnum{8} uses standard bounds on the error function together with $n \ge n^\star$ for the first integral;
        and
        \texteqnum{9} uses that $n^\star \ge c^2 / \sigma^2$.
\end{itemize}
We conclude by combining the two parts \texteqnum{5} and \texteqnum{9}.
\qed

\subsection{Proof of the \texorpdfstring{$\tmodelbonus$}{Tmodelbonus} Bound, Lemma~\ref{lem:tmodel_bonus_bound}} \label{sec:tmodel_bonus_bound}

From the definition of $\bkh$ and $\max\{a,b\} \leq a + b$ for $a,b\geq0$, we have
\begin{align*}
    \sum_{k,h} \bkh(\skh,\akh)
    &= \sum_{k,h}  b\left( P^k_{h,\skh,\akh}, \nkh(\skh,\akh), \vkhp \right) \\
    &\leq \sum_{k,h}  b_1\left( P^k_{h,\skh,\akh}, \nkh(\skh,\akh), \vkhp \right) + \sum_{k,h} b_2\left( P^k_{h,\skh,\akh}, \nkh(\skh,\akh), \vkhp \right)
\end{align*}
Now, fix $k,h$, and write $\hat P = P^k_{h,\skh,\akh}$, $P = P_{h,\skh,\akh}$, $n = \nkh(\skh,\akh)$, $\hat V = \vkhp$, and $\vstar = \vstar_{h+1}$ for ease of notation.
We again utilize the strategy of splitting $\hat V$ into $\vstar$ and $\hat V - \vstar$.
In particular, $\varp(P, \hat V) \leq 2\varp(P,\vstar) + 2\var(P,\hat V - \vstar)$.
Subsequently, under $\econc$, Lemma~\ref{lem:econc} gives us
\begin{align*}
\wtvar\left(\hat P, n, \hat V\right)
&\lesssim \varp(P, \hat V) + \frac{H^2SL}{n} \\
&\lesssim \varp(P, \vstar) + \var(P, \hat V - \vstar) + \frac{H^2SL}{n}
\end{align*}
which due to the monotonicity of $x\mapsto x\logpp\frac{B}{nx}$ implies
\begin{align*}
    b_1\left(\hat P, n, \hat V\right) 
    &\leq \sqrt{\frac{\alpha \wtvar(\hat P, n, \hat V) \logpp \frac{B}{n\wtvar( \hat P, n, \hat V)}}{n}} \\
    &\lesssim \sqrt{\frac{\varp( P, \vstar) \logpp \frac{B}{n\varp(P, \vstar)}}{n}} + \sqrt{\frac{\var( P, \hat V - \vstar) L}{n}} + \frac{ HSL}{n}.
\end{align*}
Thus, under $\econc$, we have
\begin{align*}
    \tmodelbonus
    &= \sum_{k,h} \bkh(\skh,\akh) \\
    &\lesssim \sum_{k,h}  b_1\left( P^k_{h,\skh,\akh}, \nkh(\skh,\akh), \vkhp \right) + H^3S^2AL^2 \\
    &\lesssim \sum_{k,h} \sqrt{\frac{\varp( P_{h,\skh,\akh}, \vstar_{h+1}) \logpp \frac{B}{\nkh(\skh,\akh)\varp(P_{h,\skh,\akh}, \vstar_{h+1})}}{\nkh(\skh,\akh)}} \\
    &\quad+ \sum_{k,h}\sqrt{\frac{\var( P_{h,\skh,\akh}, \vkhp - \vstar_{h+1}) L}{\nkh(\skh,\akh)}} + \sum_{k,h}\frac{ HSL}{\nkh(\skh,\akh)} + H^3S^2AL^2.
\end{align*}
We can bound the third sum by $H^2S^2AL^2$ via Lemma~\ref{lem:counts_sum}. 
Further, we can handle the second sum exactly as we handled a similar looking term while bounding $\tmodeldiff$.
Specifically, Cauchy-Schwarz gives us
\begin{align*}
    \sum_{k,h}\sqrt{\frac{\var( P_{h,\skh,\akh}, \vkhp - \vstar_{h+1}) L}{\nkh(\skh,\akh)}} 
    &\leq \sqrt{\sum_{k,h} \var(P_{h,\skh,\akh},\vkhp-\vstar_{h+1})L} \sqrt{\sum_{k,h} \frac{1}{\nkh(\skh,\akh)}}  \\
    &\leq \sqrt{HSAL^2\vardiff} .
\end{align*}
It remains to bound the first sum.
We have
\begin{align*}
    & \sum_{k,h} \sqrt{\frac{\varp( P_{h,\skh,\akh}, \vstar_{h+1}) \logpp \frac{B}{\nkh(\skh,\akh)\varp(P_{h,\skh,\akh}, \vstar_{h+1})}}{\nkh(\skh,\akh)}} \\
    &\leq H^2SAL^2 + \sum_{k,h \colon \Nkh(\skh,\akh)\geq 1} \sqrt{\frac{\varp( P_{h,\skh,\akh}, \vstar_{h+1}) \logpp \frac{B}{\nkh(\skh,\akh)\varp(P_{h,\skh,\akh}, \vstar_{h+1})}}{\nkh(\skh,\akh)}} \\
    &\leq H^2SAL^2 + \sum_{h,s,a} \sum_{n=1}^{N_h(s,a)} \sqrt{\frac{\varp( P_{h,s,a}, \vstar_{h+1}) \logpp \frac{B}{n\varp(P_{h,s,a}, \vstar_{h+1})}}{n}} \\
    &\overset{\eqnum{1}}\lesssim H^2SAL^2 + \sum_{h,s,a} \left( \sqrt{B} + \frac{\varp(P_{h,s,a},\vstar_{h+1}) N_h(s,a)}{\sqrt{B}} \right) \\
    &\overset{\eqnum{2}}= H^2SAL^2 + HSA\sqrt{B} + \frac{\varstar}{\sqrt{B}},
\end{align*}
with \texteqnum{1} following from Lemma~\ref{lem:deterministic_moss_sum} bounding the MOSS-style sum and \texteqnum{2} holding by the definition of $\varstar$.
By the choice of $B=\frac{HK}{SA}$, the first sum is therefore bounded by $\sqrt{H^3SAK} + \sqrt{\frac{SA}{HK}} \varstar + H^2SAL^2$ (up to a universal constant).

Since in the worst case $\bkh(\skh,\akh)\lesssim H^2SL$, we conclude that
\begin{align*}
    \E[\tmodelbonus]
    &= \E[\ind(\econc)\tmodelbonus] + \E[\ind(\econc^c)\tmodelbonus] \\
    &\lesssim \sqrt{H^3SAK} + H^3S^2AL^2 + \sqrt{HSAL^2\E[\vardiff]} + \sqrt{\frac{SA}{HK}} \E[\varstar] + \frac{H^3SLK}{K^3} \\
    &\lesssim \sqrt{H^3SAK} + H^3S^2AL^2 + \sqrt{HSAL^2\E[\vardiff]}.
\end{align*}
Here, the first inequality is due to Jensen's inequality and Lemma~\ref{lem:econc} bounding $\P(\econc^c) \leq 1/K^3$, and the second inequality is by Lemma~\ref{lem:var_sum} bounding $\E[\varstar]\lesssim H^2K$.
\qed

\subsection{Proof of the \texorpdfstring{$\vardiff$}{Vardiff} Bound} \label{sec:vardiff_bound_proof}

Fix arbitrary $k\in[K]$.
For each $h\in[H+1]$, write $D_h \coloneq \vkh - \vstar_h$.
We directly compute for each $h$,
\begin{align*}
    &\var(P_{h,\skh,\akh}, D_{h+1}) \\
    &= P_{h,\skh,\akh}(D_{h+1})^2 - (P_{h,\skh,\akh}D_{h+1})^2 \\
    &\leq P_{h,\skh,\akh}(D_{h+1})^2 - (D_h(\skh))^2 + (D_h(\skh))^2 - (P_{h,\skh,\akh}D_{h+1})^2 \\
    &= P_{h,\skh,\akh}(D_{h+1})^2 - (D_h(\skh))^2 + (D_h(\skh) + P_{h,\skh,\akh}D_{h+1})(D_h(\skh) - P_{h,\skh,\akh}D_{h+1}) \\
    &\leq P_{h,\skh,\akh}(D_{h+1})^2 - (D_h(\skh))^2 + 2H\left|D_h(\skh) - P_{h,\skh,\akh}D_{h+1}\right| \\
    &= P_{h,\skh,\akh}(D_{h+1})^2 - (D_h(\skh))^2 + 4H\left(D_h(\skh) - P_{h,\skh,\akh}D_{h+1}\right)_+ \\
    &\qquad\qquad\qquad\qquad\qquad\qquad\qquad\qquad\qquad - 2H\left(D_h(\skh) - P_{h,\skh,\akh}D_{h+1}\right) \\
    &\leq P_{h,\skh,\akh}(D_{h+1})^2 - (D_h(\skh))^2 + 4H\left((P^k_{h,\skh,\akh} - P_{h,\skh,\akh})\vkhp + \bkh(\skh,\akh)\right)_+ \\
    &\qquad\qquad\qquad\qquad\qquad\qquad\qquad\qquad\qquad - 2H\left(D_h(\skh) - P_{h,\skh,\akh}D_{h+1}\right).
\end{align*}
Observe that
\begin{align*}
    &\sum_h \E[P_{h,\skh,\akh}(D_{h+1})^2 - (D_h(\skh))^2] \\
    &= -\E\left[(D_1(\sko))^2\right] + \sum_h \E[P_{h,\skh,\akh}(D_{h+1})^2 - (D_{h+1}(\skhp))^2] \\
    &= -\E\left[(D_1(\sko))^2\right]
    \leq 0
\end{align*}
and
\begin{align*}
    &\sum_h \E[D_h(\skh) - P_{h,\skh,\akh}D_{h+1}] \\
    &= \E\left[D_1(\sko)\right] + \sum_h \E[D_{h+1}(\skhp) - P_{h,\skh,\akh}D_{h+1}] \\
    &= \E\left[D_1(\sko)\right]
    = \E\left[\vko(\sko) - \vstar_1(\sko)\right].
\end{align*}
Subsequently, we have
\begin{align*}
&\E\left[\sum_{h=1}^H \var(P_{h,\skh,\akh},D_{h+1})\right]\\ 
&\leq 4H\E\left[ \sum_{h=1}^H \left( (P^k_{h,\skh,\akh} - P_{h,\skh,\akh}) \vkhp + \bkh(\skh,\akh) \right)_+ \right] + 2H\E\left[(\vstar_1(\sko) - \vko(\sko))_+\right].
\end{align*}
We obtain the desired result by summing over $k$ and applying the definitions of $\topt$ and $\tmodel$.
\qed

\section{Proof of Theorem~\ref{thm:moss_conc}} \label{sec:moss_conc_proof}

In this section we prove Theorem~\ref{thm:moss_conc}.

\begin{theorem}[Restatement of Theorem~\ref{thm:moss_conc}] \label{thm:var_moss_conc}
Let $X_1,X_2,\dots$ be i.i.d.\ random variables such that
\[
\E[X_i]=\mu,
\qquad
\var(X_i)=\sigma^2,
\qquad
|X_i-\mu|\leq c
\quad\text{a.s.}
\]
For $n\geq 1$, write $\hat{\mu}_n \coloneq \frac{1}{n}\sum_{i=1}^n X_i$.
Set $\alpha_0=4$ and $\beta_0=2/3$. 
Then, for every $\delta\in(0,1)$ and $\Delta>0$,
\[
\P\left(
    \exists n\ge 1:
    \hat{\mu}_n
    +
    \sqrt{
    \frac{\alpha_0\sigma^2}{n}
    \logp\left(\frac{1}{n\delta}\right)
    }
    +
    \frac{\beta_0 c}{n}
    \logp\left(\frac{1}{n\delta}\right)
    +
    \Delta
    \leq \mu
\right)
\leq
16 \, \delta \cdot \parens*{
    \frac{\sigma^2}{\Delta^2}
    +
    \frac c{\Delta}
} \med .
\]
\end{theorem}

We will rely on the technical lemmas below.

\begin{lemma}[Maximal Bernstein inequality]
\label{lemma_maximal_bernstein}
    Let $(X_n)$ be a sequence of i.i.d.~random variables with mean $\mu$ and variance $\sigma^2$ and such that $|X_n - \mu| \le c$ a.s.
    Let $S_n := X_1 + \ldots + X_n$.
    Then, for every integer $N\geq 1$ and $x > 0$,
    \begin{equation*}
        \P \parens*{
            \max_{1 \le n \le N} (S_n - \E[S_n])
            \ge x 
        }
        \le \exp \parens*{
            \frac{-x^2}{2 (N \sigma^2 + c x / 3)}
        }
        \; .
    \end{equation*}
\end{lemma}

\begin{proof}[Proof of Lemma \ref{lemma_maximal_bernstein}]
    This result is folklore. 

    The cases $c = 0$ and $\sigma = 0$ are trivial, because in either case $X_n = \mu$ a.s.~and the probability bound is trivial.
    So, we assume in the sequel that $c > 0$ and $\sigma > 0$. 
    Following typical computations for the proof of Bernstein's inequality, the Laplace transform of $X_n$ can be bounded as such (see \cite[Corollary 2.11 and the discussion that follows it]{boucheron2013concentration}):
    \begin{equation}
        \forall \lambda \in \parens*{0, \frac 3c}:
        \wide
        \log \E [ \exp \parens*{X_k - \mu} ]
        \le
        \frac{\lambda^2 \sigma^2}{2 \parens*{1 - \frac 13 \lambda c} }
        \eqcolon 
        \psi (\lambda)
        \med .
    \end{equation}
    So for all $\lambda \in (0, 3/c)$, $W_n^\lambda \coloneq \exp \parens*{\lambda (S_n - \E[S_n]) - n \psi(\lambda)}$ is a non-negative super-martingale.
    So:
    \begin{align*}
        \P \parens*{
            \max_{1 \le n \le N} (S_n - \E[S_n])
            \ge x 
        }
        & \le 
        \inf_{\lambda > 0}
        \P \parens*{
            \max_{1 \le n \le N} 
            W_n^\lambda
            \ge 
            \ee^{- N \psi(\lambda) + \lambda x}
        }
        \\ & \overset{\eqnum{1}}\le
        \inf_{\lambda > 0}
        \braces*{
            \E [W_0^\lambda]
            \ee^{N \psi(\lambda) - \lambda x}
        }
        \le
        \exp \parens*{
            - \frac{x^2}{2 (N \sigma^2 + c x / 3)}
        }
    \end{align*}
    where 
    \texteqnum{1} follows by Ville's inequality
    and
    \texteqnum{2} is obtained by picking $\lambda = x / (N \sigma^2 + c x / 3)$.
\end{proof}

\begin{lemma}
\label{lemma_technical_function_bernstein}
    For all $\alpha, \beta > 0$, the function $\phi \equiv \phi_{\alpha, \beta} : \R_{\ge 0} \to \R_{\ge 0}$ given by
    \begin{equation*}
        \phi(x) \equiv \phi_{\alpha, \beta}(x)
        :=
        \frac{x^2}{2 (\alpha + \beta x)}
    \end{equation*}
    is super-additive\footnote{That is, $\phi(x + y) \ge \phi(x) + \phi(y)$.} and increasing on its domain.
\end{lemma}

\begin{proof}[Proof of Lemma~\ref{lemma_technical_function_bernstein}]
    The derivative of $\phi$ is $\phi'(x) = \frac{x(2 \alpha + \beta x)}{2(\alpha + \beta x)^2} > 0$ for $x > 0$, so $\phi$ is increasing on $\R_{\ge 0}$.
    For the super-additivity, remark that 
    \begin{equation*}
        \phi(x + y) - \phi(x) - \phi(y)
        =
        \frac{
            \alpha x y (2 \alpha + \beta (x + y))
        }{
            2 (\alpha + \beta x) (\alpha + \beta y) (\alpha + \beta (x + y))
        }
        \ge 0
    \end{equation*}
    for $x, y \ge 0$.
\end{proof}

\begin{proof}[Proof of Theorem~\ref{thm:var_moss_conc}]
    We write $\mathcal{E} \equiv \mathcal{E}_{\delta, \Delta}$ the event of interest.
    If $\sigma^2 = 0$ then $X_k = \mu$ a.s.~so $\P(\mathcal{E}) = 0$ trivially.
    In the sequel, we assume $\sigma^2 > 0$.
    We bound $\P (\mathcal{E})$ by doing a union bound with a geometric grid:
    \begin{align*}
        \P (\mathcal{E})
        & \le
        \sum_{\ell=0}^\infty
        \P \parens*{
            \exists n \in \llbracket 2^\ell, 2^{\ell+1} \llbracket
            :
            \hat{\mu}_n 
            + \sqrt{ \frac{4 \sigma^2}n \logp \parens*{\frac 1{n\delta}}}
            + \frac{2c}{3n} \logp \parens*{\frac 1{n\delta}}
            + \Delta
            \le \mu
        }
        \\ & \overset{\eqnum{1}}\le
        \sum_{\ell=0}^\infty
        \P \parens*{
            \exists n \le 2^{\ell+1}
            :
            S_n - n \mu
            + \sqrt{ 4 \sigma^2 2^{\ell} \logp \parens*{\frac 1{2^{\ell+1} \delta}}}
            + \frac{2c}3 \logp \parens*{\frac 1{2^{\ell+1}\delta}}
            + 2^\ell \Delta
            \le 0
        }
        \\ & \overset{\eqnum{2}}\le
        \sum_{\ell=0}^\infty
        \exp \parens*{
            - \phi_\ell \parens*{
                \sqrt{ 4 \sigma^2 2^{\ell} \logp \parens*{\frac 1{2^{\ell+1} \delta}}}
                + \frac{2c}3 \logp \parens*{\frac 1{2^{\ell+1}\delta}}
                + 2^\ell \Delta
            }
        }
        \\ & \overset{\eqnum{3}}\le
        \sum_{\ell=0}^\infty
        \underbrace{
            \exp \parens*{
                - \phi_\ell \parens*{
                    \sqrt{ 4 \sigma^2 2^{\ell} \logp \parens*{\frac 1{2^{\ell+1} \delta}}}
                    + \frac{2c}3 \logp \parens*{\frac 1{2^{\ell+1}\delta}}
                }
            }
        }_{\text{Factor } \mathrm{(I)}_\ell}
        \cdot 
        \underbrace{
            \exp \parens*{
                - \phi_{\ell} \parens*{
                    2^\ell \Delta
                }
            }
            \vphantom{
                \exp \parens*{
                    - \phi_\ell \parens*{
                        \sqrt{ 4 \sigma^2 2^{\ell} \logp \parens*{\frac 1{2^{\ell+1} \delta}}}
                        + \frac{2c}3 \logp \parens*{\frac 1{2^{\ell+1}\delta}}
                    }
                }
            }
        }_{\text{Factor } \mathrm{(II)}_\ell}
    \end{align*}
    where 
    in \texteqnum{1} we use that $n \mapsto \logp ( \frac 1{n\delta})$ is non-increasing;
    in \texteqnum{2} we invoke the maximal Bernstein inequality (Lemma~\ref{lemma_maximal_bernstein}) and introduce the notation $\phi_\ell \equiv \phi_{\alpha_\ell, \beta_\ell}$ (see Lemma~\ref{lemma_technical_function_bernstein}) with $\alpha_\ell := 2^{\ell + 1} \sigma^2$ and $\beta_\ell \equiv \beta := c / 3$;
    and
    in \texteqnum{3} we use the super-additivity of $\phi_\ell$ (see Lemma~\ref{lemma_technical_function_bernstein}).
    
    We proceed by bounding the two factors $(\mathrm{I})_\ell$ and $(\mathrm{II})_\ell$.

    Fix $\ell \ge 0$.
    For short, write $m := 2^\ell$ and $L := \logp (\frac 1{2 m \delta})$. 
    Recall that $\alpha \equiv \alpha_\ell = 2 m \sigma^2$ and $\beta = c/3$.

    \begin{itemize}
        \item
        \emph{Factor $\mathrm{(I)}_\ell$:}
        For the first factor $\mathrm{(I)}_\ell$, by setting $q := \sqrt{2 \alpha L} + 2 \beta L$, we show via straight-forward algebra that $q^2 - 2 L (\alpha + \beta q) = 2 \beta L \sqrt{2 \alpha L} \ge 0$.
        This readily implies $\smash{\phi_\ell (q) = \frac{q^2}{2 (\alpha + \beta q)} \ge L}$.
        So:
        \begin{align}
        \nonumber
            \mathrm{(I)}_\ell
            \equiv
            \exp \parens*{
                - \phi_\ell \parens*{
                    - \sqrt{2 \alpha L}
                    + 2 \beta L
                }
            }
            & \equiv 
            \exp \parens*{ - \phi_\ell (q) }
            \\ & \le 
        \label{equation_variance_aware_moss_1}
            \exp \parens*{ 
                - \logp \parens*{ \frac 1{2 m \delta} }
            }
            \overset{\eqnum{4}}\le
            2 m \delta
        \end{align}
        where 
        \texteqnum{4} follows from $\exp (-\logp(1/x)) \le \min \braces{1, x} \le x$.

        \item 
        \emph{Factor $\mathrm{(II)}_\ell$:}
        Continuing with the second factor $\mathrm{(II)}_\ell$, we have
        \begin{align}
        \label{equation_variance_aware_moss_2}
            \mathrm{(II)}_\ell
            \equiv 
            \exp \parens*{
                - \frac{
                    m^2 \Delta^2
                }{
                    2 \parens*{
                        2 m \sigma^2 + 
                        \frac {c m \Delta}3
                    }
                }
            }
            = \exp \parens*{
                - \frac{
                    m \Delta^2
                }{
                    4 \sigma^2 + \frac {2 c \Delta}3
                }
            }
            \; .
        \end{align}
    \end{itemize}
    Injecting Eq.~\eqref{equation_variance_aware_moss_1} and Eq.~\eqref{equation_variance_aware_moss_2} back in to \texteqnum{3}, we pursue the bound of $\P(\mathcal{E})$ as such:
    \begin{align*}
        \P (\mathcal{E})
        & \overset{\eqnum{5}}\le
        2 \delta 
        \sum_{\ell=0}^\infty
        2^\ell 
        \exp \parens*{
            - 2^\ell 
            \cdot \frac{ \Delta^2 }{ 4 \sigma^2 + \frac{2c \Delta}3 } 
        }
        \\ & \overset{\eqnum{6}}\le
        2 \delta 
        \cdot 
        2 \integral_{0}^\infty
        \exp \parens*{
            - \frac{ \Delta^2 }{ 4 \sigma^2 + \frac{2c \Delta}3 } 
            \cdot x
        }
        \mathrm{d} x
        = 
        \frac {4\delta}{\Delta^2} \parens*{
            4 \sigma^2 + \frac{2 c \Delta}3
        }
        \leq 
        16\, \delta \parens*{
            \frac{\sigma^2}{\Delta^2} + \frac{c}{\Delta}
        }.
    \end{align*}
    where
    \texteqnum{5} follows by injecting Eq.~\eqref{equation_variance_aware_moss_1} and Eq.~\eqref{equation_variance_aware_moss_2} into \texteqnum{3};
    and
    \texteqnum{6} follows by an integral-series comparison.
\end{proof}

\section{Best Policy Identification}
\label{sec:BPI_appendix}

\subsection{Related Work}
First we discuss additional related work on the problem of best-policy identification in episodic MDPs.

\citet{jin2018qlearning} describe a standard regret-to-PAC conversion for episodic MDPs which selects a policy from a single uniformly random episode, and only achieves a constant success probability (once the average regret is sufficiently small).
A regret-to-PAC conversion from a high-probability regret bound can instead retain the same success probability if the conversion instead outputs a mixture policy, but it is generally more common to require the learner to output a single policy \citep{wagenmaker2022beyond}.
\citet{menard2021fast} also discuss the poor dependence on the failure probability if a single uniformly random episode's policy is chosen.
However, \cite{kaufmann2021adaptive} note that it is instead possible to amplify the success probability by estimating the values of multiple policies that have constant probability of being $O(\varepsilon)$-optimal and outputting the best; this is the same approach used within our Algorithm \ref{alg:BPI_alg}.
The minimax sample complexity lower bound $\Omega(H^3SA \log(1/\delta)/\varepsilon^2)$ (measured in episodes) is due to \cite{domingues2021episodic}, and it requires the mild additional assumption that $H \geq \Omega(\log_A S)$, as well as that $S, A$, and $H/\varepsilon$ are at least certain absolute constants. We refer to their Theorems 7 and 10 for the precise requirements.
This lower bound allows the learner to have a random stopping time and concerns its expected stopping time, so it applies to our Corollary \ref{cor:pac_bound} which has a deterministic stopping time. Some prior algorithms discussed below use random stopping times and only bound them with high probability, but such results can be easily converted to algorithms with deterministic stopping times by simply terminating the algorithm once the budget from its proven complexity guarantee is reached.

Next we compare with other sample complexity bounds for $(\varepsilon, \delta)$-PAC best-policy identification, measuring episodes and assuming rewards are in $[0,1]$. Throughout we abbreviate $L_1 = \log(eSAH/\delta), L_2 = \log(e SAH^2 L_1/\varepsilon)$.
The \texttt{BPI-UCRL} algorithm of \citet{kaufmann2021adaptive} has a suboptimal outer $H^4$ horizon dependence rather than $H^3$.
The \texttt{BPI-UCBVI} algorithm of \citet{menard2021fast} give a sample complexity bound for sufficiently small $\varepsilon$ of $O \left( \frac{H^3SA}{\varepsilon^2}L_1 L_2^2 \right)$.
After expanding the hidden logarithmic factors, to the best of our knowledge the \texttt{ORLC} algorithm of \cite{dann_policy_2019} yields a PAC sample complexity bound of $O\left( \left(\frac{H^3 SA}{\varepsilon^2} + \frac{H^5 S^2 A}{\varepsilon}  \right) L_2 (L_1 + \log L_2)\right)$. After augmenting the state space with a time index to apply the time-homogeneous results of \citet{lee2025minimax} to our inhomogenous setting, their \texttt{EQO}-based method achieves a PAC complexity bound of $O\left( \frac{H^3 SA}{\varepsilon^2}L_1 L_2 + \frac{H^3 S^2 A}{\varepsilon} (L_1 + \log L_2)L_2 \right)$. 
In summary, the best sample complexity guarantees for best-policy identification still suffer from a gap of multiplicative logarithmic factors relative to the minimax lower bound.

Additionally, the aforementioned regret-to-PAC conversions \citep{kaufmann2021adaptive} could be applied to any algorithms that enjoy suitable sublinear expected regret or high-probability regret bounds, including the nearly minimax-optimal algorithms from Table \ref{table:minimax_optimal_algorithms}, but such reductions would still retain the suboptimal logarithmic factors. (We note that if there did exist a high-probability regret bound with suboptimal dependence only on $1/\delta$, then, because the regret-to-PAC reduction of \citet{kaufmann2021adaptive} only requires constant success probability, the algorithm could still be used to obtain an optimal sample complexity for best-policy identification in some regimes. However, to the best of our knowledge even for $\frac{1}{\delta} = O(1)$ all previous work still contains suboptimal logarithmic factors in $S$, $A$, $H$, or $K$, so this is not possible from existing guarantees.) Of particular interest is the result of \citet{zhang2025settling}, because despite the suboptimal multiplicative logarithmic factors in its leading order $\sqrt{K}$ regret term, it does not suffer from additional additive ``burn-in'' regret terms, and so combining with the regret-to-PAC conversion of \citet{kaufmann2021adaptive} would yield a best-policy identification sample complexity of $O \left( \frac{H^3 SA L_\delta}{\varepsilon^2} \left(\log(eSAH/\varepsilon) \right)^5 + \frac{H^2 L_\delta^2}{\varepsilon^2} \right)$ (recall $L_\delta = \log(2/\delta)$), which can be superior to our Corollary \ref{cor:pac_bound} in regimes of sufficiently large $\varepsilon$.

\cite{wagenmaker2022beyond} and \cite{tirinzoni_optimistic_2023} study instance-dependent complexity of best-policy identification, with dependencies on suboptimality gaps and state visitation probabilities, rather than the minimax complexity for fixed $S,A,H, \varepsilon, \delta$. \citet{kone_optimal_2026} study the complexity of exact identification of an optimal policy in the regime of small $\delta$.

\subsection{Algorithm and Proof of Corollary \ref{cor:pac_bound}}
Next we present the algorithm which is the subject of Corollary \ref{cor:pac_bound}, as well as the proof.
In Algorithm \ref{alg:BPI_alg}, $\CBPI, \CBPI'$ are absolute constants which are defined in the proof below in terms of the constants for which the regret bound in Theorem \ref{thm:minimax_regret} holds.

\begin{algorithm}[H]
\caption{\texttt{MORE-BPI}} 
\label{alg:BPI_alg}
\SetAlgoLined
\DontPrintSemicolon

$m= \lceil \log_2 (\frac{2}{\delta}) \rceil, K_0 = \left\lceil\frac{\CBPI SAH^3}{\varepsilon^2} + \CBPI' H^3S^3 A (\log(eHSA))^4 \right\rceil, M = \lceil \frac{9 H^2 \log(4m/\delta)}{2 \varepsilon^2} \rceil $ \;

\For{$i=1,\dots, m$}{
    Independently run $K_0$ episodes of Algorithm \ref{alg:main_alg}, generating policies $\pi^{(i), 1}, \dots, \pi^{(i), K_0}$  \;
    Independently sample $J_i \sim \textrm{Uniform}\{1, \dots, K_0\}$ and let $\pimix^{(i)} = \pi^{(i), J_i}$ \;
}
\For(\tcp*[f]{estimate the value of policy $\pimix^{(i)}$}){$i=1,\dots, m$}{
    Follow policy $\pimix^{(i)}$ for $M$ episodes and record cumulative returns within each episode $R_{i, 1}, \dots, R_{i, M}$ \;
    $\widehat{V}^i \gets \frac{1}{M}\sum_{j=1}^M R_{i,j} $ \;
}
Let $I \in \argmax_{i} \widehat{V}^i$ and return $\pimix^{(I)}$\;
\end{algorithm}

Algorithm \ref{alg:BPI_alg} has two phases. First it generates $m$ candidate policies $\pimix^{(1)}, \dots, \pimix^{(m)}$. For each $i \in [m]$ this is done by running $K_0$ episodes of Algorithm \ref{alg:main_alg}, where $K_0$ is chosen so that the expected average regret over $K_0$ episodes is at most $\varepsilon/6$. As shown in the proof, this implies via Markov's inequality that if we sample an episode $J_i$ uniformly at random from $[K_0]$ and use the policy $\pi^{(i), J_i}$ from that episode, then $\pi^{(i), J_i}$ (which is the definition of $\pimix^{(i)}$) will be $\varepsilon/3$-optimal with constant probability. 

The success probability is then amplified by estimating the value of all $m = O(\log_2(2/\delta))$ policies and choosing the (estimated) best one. This amplification step is standard and is basically identical to \cite[Proposition 12]{kaufmann2021adaptive}, but we provide the short argument for completeness.

\begin{lemma}
    \label{lem:BPI_n_to_eps_conversion}
    Let $X > 1$. If $K \geq e^4 X (\log(e^4X))^4$, then $\frac{K}{(\log K)^4} \geq X$.
\end{lemma}
\begin{proof}
    Define the function $g(x) = \frac{x}{(\log x)^4}$. This function is non-decreasing on $[e^4, \infty)$ because
    \begin{equation*}
        g'(x) = \frac{1}{(\log x)^4} \left(1 - \frac{4}{\log x} \right) \geq 0.
    \end{equation*}
    Letting $x_0 = e^4 X (\log(e^4X))^4 \geq e^4$, since $x_0 \geq e^4$ it suffices to show that $g(x_0) \geq X$, because then for all $K \geq x_0$ we have
    \begin{equation*}
        \frac{K}{(\log K)^4} = g(K) \geq g(x_0) \geq X.
    \end{equation*}
    To do so, we calculate
    \begin{equation*}
        g(x_0) = \frac{e^4 X (\log(e^4X))^4}{(\log x_0)^4} \geq \frac{e^4 X (\log(e^4X))^4}{e^4 (\log(e^4X))^4} = X
    \end{equation*}
    where the inequality step comes from the bound
    \begin{equation*}
        \log x_0 = \log (e^4 X) + 4 \log \log (e^4X) \leq \log (e^4 X) + \frac{4}{e} \log (e^4X) \leq e \log (e^4 X)
    \end{equation*}
    using the standard fact that $\log t \leq \frac{t}{e}$ (for $t > 0$).
\end{proof}

\begin{proof}[Proof of Corollary \ref{cor:pac_bound}]
From Theorem \ref{thm:minimax_regret}, there exist $C_1, C_2 \geq 1$ such that
\begin{equation*}
    \E [\reg(K)] \leq \sqrt{C_1 H^3 SAK} + C_2 H^3 S^2 A (\log(K))^2.
\end{equation*}
For the first term to dominate, we have the implications
\begin{align*}
    & \sqrt{C_1 H^3 SAK} \geq C_2 H^3 S^2 A (\log(K))^2 \\
    \iff & K \geq \frac{ C_2^2H^6 S^4 A^2 (\log(K))^4}{C_1H^3 SA} = \frac{C_2^2}{C_1} H^3 S^3 A (\log(K))^4 \\
    \impliedby & K \geq e^4 \frac{C_2^2}{C_1} H^3 S^3 A \left( \log \left( e^4\frac{C_2^2}{C_1} H^3 S^3 A \right) \right)^4 \\
    \impliedby & K \geq C_4 H^3S^3 A (\log(eHSA))^4
\end{align*}
where the penultimate implication is by Lemma \ref{lem:BPI_n_to_eps_conversion}, and the final implication is by choosing a sufficiently large constant $C_4$.
Hence as long as $K \geq C_4 H^3S^3 A (\log(eHSA))^4$, we have
\begin{equation*}
    \E [\reg(K)] \leq 2\sqrt{C_1 H^3 SAK} = \sqrt{C_3 H^3 SAK}
\end{equation*}
where $C_3 = 4 C_1$.

Next, for any $i \in [m]$, the expected suboptimality of the candidate policy $\pimix^{(i)} = \pi^{(i), J_i}$ is bounded as
\begin{equation*}
    \E\left[ \vstar_1(\mu) - V_1^{\pimix^{(i)}}(\mu) \right] = \E\left[\frac{1}{K_0} \sum_{k=1}^{K_0} \left(\vstar_1(\mu) - V_1^{\pi^{(i),k}}(\mu)\right) \right] = \frac{\E[\reg(K_0)]}{K_0} \leq \frac{ \sqrt{C_3 H^3 SAK_0}}{K_0} = \sqrt{\frac{C_3 H^3 SA}{K_0}}.
\end{equation*}
Here the first equality is because the conditional expectation (on the transitions used in forming the policies $\pi^{(i),1}, \dots, \pi^{(i), K_0}$) of $V_1^{\pimix^{(i)}}(\mu)$ is equal to $\frac{1}{K_0}\sum_{k=1}^{K_0} V_1^{\pi^{(i),k}}(\mu)$. The second equality uses the fact that each $\pi^{(i),k}$ is chosen before $s_1^k$ is sampled from $\mu$, so $\E\left[\vstar_1(\mu) - V_1^{\pi^{(i),k}}(\mu) \right] = \E\left[\vstar_1(s_1^k) - V_1^{\pi^{(i),k}}(s_1^k) \right]$.
By Markov's inequality (since $\vstar_1(\mu) - V_1^{\pimix^{(i)}}(\mu)  \geq 0$ almost surely),
\begin{equation*}
    \P\left( \vstar_1(\mu) - V_1^{\pimix^{(i)}}(\mu)  \geq \frac{\varepsilon}{3}\right) \leq \frac{\E\left[ \vstar_1(\mu) - V_1^{\pimix^{(i)}}(\mu) \right]}{\varepsilon / 3} \leq \sqrt{\frac{9 C_3 H^3 SA}{K_0 \varepsilon^2}} \leq \frac{1}{2}
\end{equation*}
where the final inequality holds if $K_0 \geq \frac{36 C_3SAH^3}{\varepsilon^2}$, which is clearly true for the choice
\begin{equation*}
    K_0 = \left\lceil\frac{36 C_3SAH^3}{\varepsilon^2} + C_4 H^3S^3 A (\log(eHSA))^4 \right\rceil,
\end{equation*}
and so we let $\CBPI = 36 C_3$ and $ \CBPI' = C_4$.

Consequently, applying this fact to the $\pimix^{(1)}, \dots, \pimix^{(m)}$ generated by Algorithm \ref{alg:BPI_alg}, and noting their independence, we can bound the probability of all these policies being $\varepsilon/3$-suboptimal as
\begin{equation*}
    \P \left( \forall i, ~ \vstar_1(\mu) - V_1^{\pimix^{(i)}}(\mu)  \geq \frac{\varepsilon}{3} \right) \leq 2^{-m} \leq \frac{\delta}{2}.
\end{equation*}
Letting $I^\star \in \argmax_i V_1^{\pimix^{(i)}}(\mu)$, we consequently have
\begin{equation*}
    \P \left( V_1^{\pimix^{(I^\star)}}(\mu) \geq  \vstar_1(\mu) - \frac{\varepsilon}{3} \right) \geq 1 -\frac{\delta}{2}.
\end{equation*}

For each fixed $i \in [m], j \in [M]$, letting $\mathcal{D}$ be the sigma-algebra generated by the candidate policies $\pimix^{(1)}, \dots, \pimix^{(m)}$, since $\E[R_{i,j} \mid \mathcal{D}] = V_1^{\pimix^{(i)}}(\mu)$ and $0 \leq R_{i,j}  \leq H$ almost surely, by (conditional) Hoeffding's inequality (using conditional independence from the data used to construct the $\pimix^{(i)}$) and the union bound we have that
\begin{equation*}
    \P \left(\forall i, ~\left|\widehat{V}^{i} -  V_1^{\pimix^{(i)}}(\mu) \right| \leq \frac{\varepsilon}{3} \right) \geq 1 - \frac{\delta}{2}
\end{equation*}
by the choice of $M = \frac{9H^2\log(4m/\delta)}{2\varepsilon^2}$.

On the intersection of both of the above good events, which has probability at least $1 - \delta$,
we have that
\begin{equation*}
    V_1^{\pimix^{(I)}}(\mu) \geq \widehat{V}^{I} - \frac{\varepsilon}{3} \geq \widehat{V}^{I^\star} - \frac{\varepsilon}{3} \geq V_1^{\pimix^{(I^\star)}}(\mu) - \frac{2 \varepsilon}{3} \geq \vstar_1(\mu) - \varepsilon.
\end{equation*}

Finally, we note that for all $\delta \in (0,1)$, from the standard bound $\log(2/\delta) \leq \frac{2}{\delta}-1$ we have that $\log(2/\delta) \leq \frac{2}{\delta}$, so $M = O\left( \frac{H^2 \log(2/\delta)}{\varepsilon^2} \right)$. The bound on the number of episodes is thus $m K_0 + m M \leq O\left( \frac{H^3 SA L_\delta}{\varepsilon^2} + \frac{H^2 L_\delta^2}{\varepsilon^2} + H^3S^3 A (\log(eHSA))^4 L_\delta \right)$ where $L_\delta = \log(2/\delta)$.
\end{proof}

\end{document}